\documentclass[twocol]{ametsocV6.1}

\title{A Physics-Constrained Neural Differential Equation Framework \\ for Data-Driven Snowpack Simulation}

\authors{Andrew Charbonneau\aff{a}\correspondingauthor{Andrew Charbonneau, acharbon@caltech.edu},
Katherine Deck,\aff{a} 
Tapio Schneider,\aff{a}
}

\affiliation{\aff{a}{California Institute of Technology}\\
}

\abstract{
This paper presents a physics-constrained neural differential equation framework for parameterization, and employs it to model the time evolution of seasonal snow depth given hydrometeorological forcings. When trained on data from multiple SNOTEL sites, the parameterization predicts daily snow depth with under 9\% median error and Nash Sutcliffe Efficiencies over 0.94 across a wide variety of snow climates. The parameterization also generalizes to new sites not seen during training, which is not often true for calibrated snow models. Requiring the parameterization to predict snow water equivalent in addition to snow depth only increases error to $\sim$12\%. The structure of the approach guarantees the satisfaction of physical constraints, enables these constraints during model training, and allows modeling at different temporal resolutions without additional retraining of the parameterization. These benefits hold potential in climate modeling, and could extend to other dynamical systems with physical constraints. }

\begin{document}

\maketitle

%
%
%
%
%
%

%
\section[Rationale]{Introduction}
Seasonal snowpacks help regulate the Earth's energy balance, provide freshwater storage, and are crucial for understanding Earth's climate. They hold economic and ecological significance, supplying a majority of the western United States' (and a sixth of the world’s) water supply and influencing agriculture, flood, drought, and avalanche hazards \citep{bulk-density_models, china-late-pred}. Their seasonal importance and susceptibility to climate change emphasize the need for ongoing modeling and monitoring on both seasonal and multi-decadal timescales.

Modeling the evolution of seasonal snow for regional or global climate applications offers a challenging problem of scales; it is the bulk properties of the snow (albedo, snow cover fraction, snow temperature, and snow water content) that are critical, yet microphysical and location-specific processes control these properties and must be taken into account. The most detailed models represent vertically-resolved snowpacks, including liquid percolation, phase changes, metamorphism effects, and other types of compaction; they are often calibrated and used on the site-level (e.g., \citealt{de2013investigating}). Models used in global climate simulations range in complexity from single-layer/bulk models to multi-layer models with parameterizations for bulk properties that are calibrated with observational data \citep{MIP_rebuttal}; the horizontal resolution of these models is typically ${\sim}10$--$100$~km where microscopic processes cannot be tractably resolved. While the laws of physics ultimately govern the evolution of snowpacks, uncertainty in how to relate essential but unresolved small-scale processes to snowpack bulk properties on the spatial scales of global models makes developing snow models a challenging task \citep{PNAS, afghanistan-nn}. This challenge is exacerbated by data availability \citep{MIP_rebuttal, cmip6-swe}.

Among bulk variables in global snow models, the snow water equivalent, $\mathrm{SWE}$ (units of $\mathrm{m}$), represents the total water storage in snow. It relates to the snow depth $z$ (units of $\mathrm{m}$) through the bulk snow density $\rho_{\mathrm{snow}}$ (units of $\mathrm{kg~m^{-3}}$) and the density of liquid water $\rho_{\mathrm{water}}$ (1000 $\mathrm{kg~m^{-3}}$) as
\begin{equation}\label{eq:swe_def}
\mathrm{SWE} = ( \rho_{\mathrm{snow}} / \rho_{\mathrm{water}} ) \; z;
\end{equation}
it is typically used as a prognostic variable in bulk models for determining snowpack mass balance. Density and depth are critical in climate models for determining the snowpack energy balance, as they influence thermal, mechanical, and optical properties, impacting mass/energy fluxes, water retention, and spring thaw \citep{cmip6-swe, density2prop}. However, for a given $\mathrm{SWE}$, the snow depth and bulk density can vary considerably over time at a single location, or between locations under similar forcings, due to compaction, melt/refreeze cycles, and changes in the density of falling snow. These variations give rise to ongoing challenges in snow modeling. 

Many prevalent snow models rely on the seminal parameterizations of \cite{kojima} and \cite{anderson} to derive snow density and depth from modeled $\mathrm{SWE}$. These works modeled compaction and microstructure metamorphism by assuming a linear relationship between the strain rate of compaction and the weight of the overlying snow and suggested empirical parameterizations that were calibrated from a select number of observational sites and laboratory experiments and analytically extended. Snow evolution in models such as Snow17, iSnobal, Noah/NoahMP, CLM5, HTESSEL, and others \citep{Snow17, PySnobal, Noah-MP, CLM5, HTESSEL, MIP_rebuttal} employ this simplified formulation based on small-scale representations; thus, it is integral to most current US and European global snow predictions.

However, these parameterizations were primarily developed for the hydrological community instead of earth system models, and their calibration to prioritize accurate $\mathrm{SWE}$ and subsequent local runoff induces a trade-off in density/depth errors. Such errors impact the snowpack energy balance, which is as crucial as mass balance in global climate simulations \citep{snow-precip-feedback1, snow-precip-feedback2}, and snow depth errors remain problematic in climate predictions \citep{MIP_rebuttal}. The inadequacy for energy tracking in these formulations has been recognized for decades and spurred development of more detailed and realistic alternatives such as CROCUS or SNOWPACK, which evolve the metamorphism of snow microstructure \citep{crocus2012-err, SNOWPACK, crocus1989}. While these complex models address the limitations of Anderson's and Kojima's original formulations, they require site-specific calibration and struggle with computational scalability for global applications. This necessitates more efficient representations for global models that can accurately predict snow depth/density. With the push toward even finer localized (1–10 km) land modeling, simple yet accurate models are increasingly essential for computational efficiency \citep{Clark, schar, Ban2021}.

Advances in computing and sensing technology have led to initiatives in data assimilation and machine learning (ML) aimed at disrupting longstanding parameterizations. Today, extensive snow depth observations surpass the resolution and precision of $\mathrm{SWE}$ and density data/estimates \citep{prevalent-z1, prev-z2}. Improving snow models with these resources has become a dominant avenue of hydrology research, with a bias towards SWE modeling, like incorporating remotely sensed depth into iSnobal to infer $\mathrm{SWE}$ \citep{iSnobal}. Several ML models (e.g., \citealt{afghanistan-nn, nunavut_ML_err, duan, Steele}) predict $\mathrm{SWE}$ or depth from meteorological and topographical inputs in specific regions. Such models show satisfactory snowpack estimation but frequently yield errors over 15\% when tested, especially beyond their training or calibration locations \citep{nunavut_ML_err, ski-resort-err, Crocus2020-err}. The ability of these empirical models to generalize to new locations or future climates and act as a universal model is limited, and their statistical or black-box nature does not inherently respect physical constraints, impeding their integration into land models \citep{bulk-density_models, china-late-pred}. Combining depth observations and ML techniques to improve depth/density parameterizations has the potential to lead to improved simulation of  key variables, benefiting global climate modeling. Capitalizing on this opportunity demands a representation that can generalize to many snowpacks and integrate with existing large-scale models. 

This work presents a novel hybrid approach to parameterization, combining physical principles with empirical modeling that structurally guarantees compliance with prescribed bounds (for example, physical consistency or conservation conditions). We showcase its utility in designing an alternative parameterization for snow depth, with ramifications for global climate and seasonal simulations. We also create a quality-controlled dataset for snow modeling and make it publicly available. Learning physically-informed representations from observational data across many locations enables robust performance that can generalize to new locations without recalibration. The customizability of the approach permits straightforward adaptation to different operational requirements and constraints, demonstrating additional capabilities with minimal adjustment. This offers a flexible, efficient, and scalable framework that is adaptable as the field evolves. The proposed approach exhibits a versatile means for enforcing (or learning) any function-based threshold on an optimizable model without modification of the training metrics, which can contribute to contemporary global snow modeling as well as other physics/ML hybrid models. 

\section{Methodology}
\subsection{Overview}
Our model choices leverage ML for seasonal snow simulation in climate models, prioritizing generalizability and computational efficiency. Contemporary paradigms in hydrology research focus on parameterizing SWE from $z$ and other data to constrain its value over global grids. However, within global climate models, SWE evolution is already well-constrained by explicitly implemented physical laws enforcing mass conservation and relatively well-understood fluxes such as sublimation, precipitation, and melt. The subsequent conversion of simulated SWE to variables like $z$ or $\rho_{\mathrm{snow}}$ is typically left to longstanding parameterizations.
These parameterizations can be pre-calibrated offline prior to use in a snow model, or are often further tuned “online” within a snow model. The data used for calibration can be indirect (non-SWE data), such as energy and water fluxes or land surface albedo, or be direct measurements of snow depth or density, or even gridded SWE estimates. The resulting global simulations are sometimes employed in refining gridded SWE for calibrating other models, creating circular estimation and biases. 

Given the relative abundance of observations of snow depth $z$ alongside additional snow variables, this instead justifies parameterizing $z$ (from physics-constrained SWE) as an alternative to established formulations to address these limitations, so that tighter relationships can be determined and evaluated on the basis of direct, high-quality data. By using primary observational data instead of assimilated/reanalysis data, this approach can avoid biases and inaccuracies, ensuring more faithful representations of the underlying processes.


We model the rate of change in snowpack height (units of $\mathrm{m~s^{-1}}$) by an ordinary differential equation (ODE) represented by an artificial neural network $M$: 
\begin{equation}\label{eq:overview}
    \frac{dz}{dt} = M\left(z, \mathrm{SWE}, \varphi, R, v, T_{\mathrm{air}}, P_{\mathrm{snow}}\right),
\end{equation}
where $\mathrm{SWE}$ is the snow-water equivalent (m), $\varphi$ is the relative humidity (between 0 and 1, the used data are measured with respect to liquid water), $R$ is the broadband solar radiative energy flux ($\mathrm{W~m^{-2}}$), $v$ is the wind speed ($\mathrm{m~s^{-1}}$), $T_{\mathrm{air}}$ is the air temperature ($^\circ \mathrm{C}$), and $P_{\mathrm{snow}}$ is the liquid water-equivalent rate of snowfall ($\mathrm{m~s^{-1}}$).  The 1D column approach permits application over any spatial grid. The chosen input variables only indirectly encode location and time dependencies through the environmental input variables, allowing the model to function in areas where topographical or temporal information is unavailable. This choice aims to enable learning about universal linear and nonlinear physical processes that apply independent of time, season, and location. Using a feed-forward neural network dependent only on the current system state aligns with land-surface models, as it matches the differential equation format used for other variables. This model is also adaptable for different applications or when $\mathrm{SWE}$ is unavailable (see sections 2\ref{s:structure} and 3\ref{s:training}).

\subsection{Model Structure} \label{s:structure}\label{s:constraints}
\begin{figure}[ht!]
        \centerline{\includegraphics[width = 15pc]{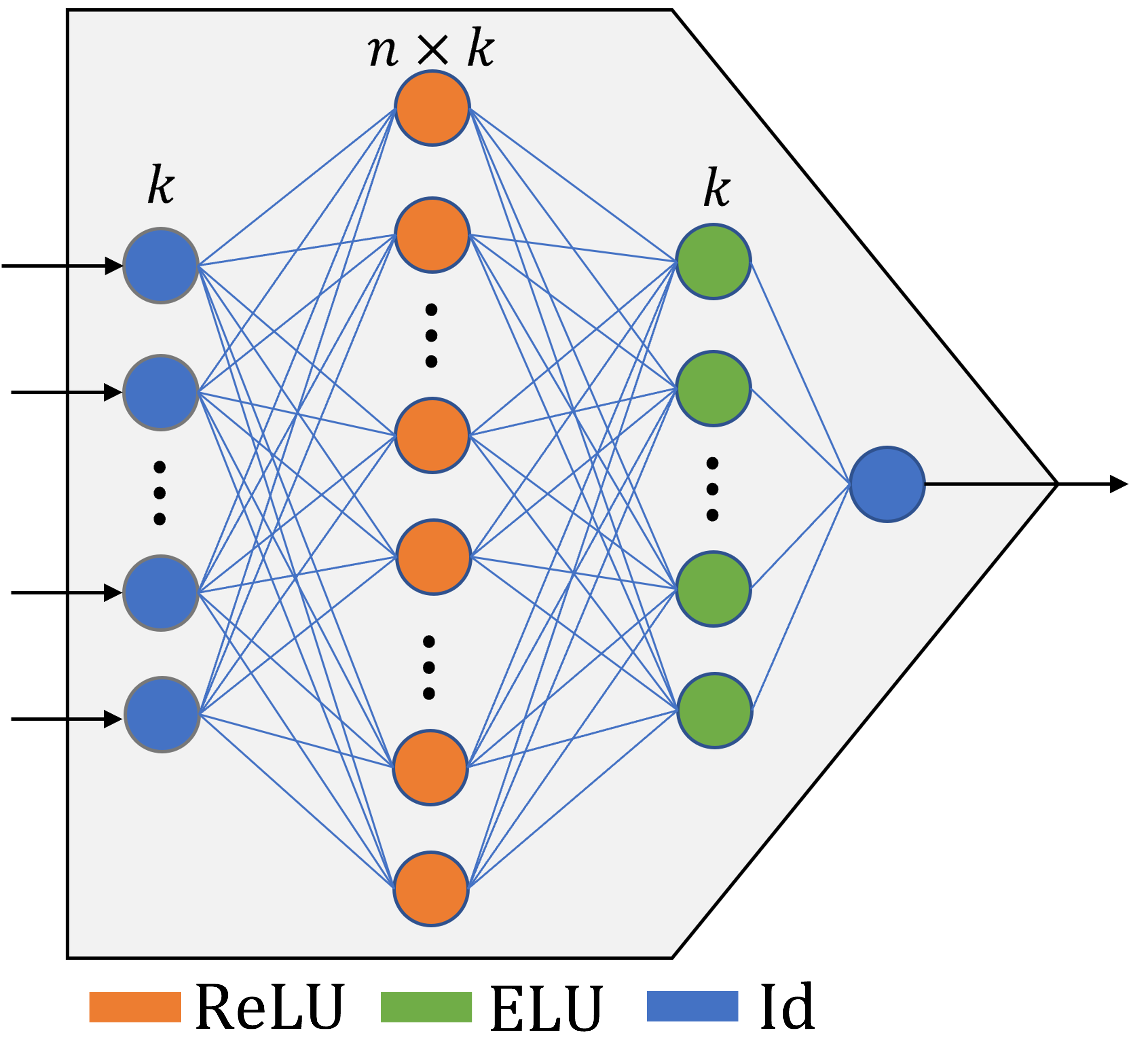}}
        \caption{Structure of the model's predictive component. Blue lines indicate a trainable linear transformation of the input (of $k$ scalar variables), including a bias. Colors indicate the activation function used upon collection at the node (ReLU = Rectified Linear Unit, ELU = Exponential Linear Unit, Id = Identity), as noted in the legend. The hyperparameter $n$ sets the width of the internal mixing layer.}
        \label{f:pred}
\end{figure}

The model $M$ consists of two components. The first is a ``predictive'' network with trainable weights to generate a $dz/dt$ prediction (Fig.\ \ref{f:pred}). For computational simplicity, only two hidden layers were used, which can also be interpreted as a regression on once-transformed features, with the transformational layer width set by the hyperparameter $n$ scaled by the number of input features $k$ (see Fig.\ \ref{f:pred}). Inputs are easily exchangeable for alternative use cases or target predictions. 

The second component consists of fixed-weight dense layers with Rectified Linear Unit (ReLU) activation, designed to impose explicit (``hard") constraints on the predictive model. This approach allows for enforcing min/max thresholds on any predictive model, without introducing penalties into the calibration metrics. Although more advanced methods exist in literature or modern coding packages \citep{enforce1, enforce2, enforce3}, our simplistic approach offers multiple advantages. Primarily, constraints are applied throughout training, leading to better gradient updates within prescribed limits (see Table \ref{t:performance} for comparison). This flexible framework supports most functions or specialized constructions, including ``learned" constraints, and can be scalably implemented in environments with minimal to no ML support, offering potential for many fields (for more on this process, see Appendix A). By guaranteeing physical constraints, it ensures stability during time-stepping and is conducive for integration into larger models without violating prescribed bounds such as conservation or consistency equations.  

\subsection{Threshold Constraints for Snowpack Prediction}

\begin{figure}[ht!]
        \centerline{\includegraphics[width = 19pc]{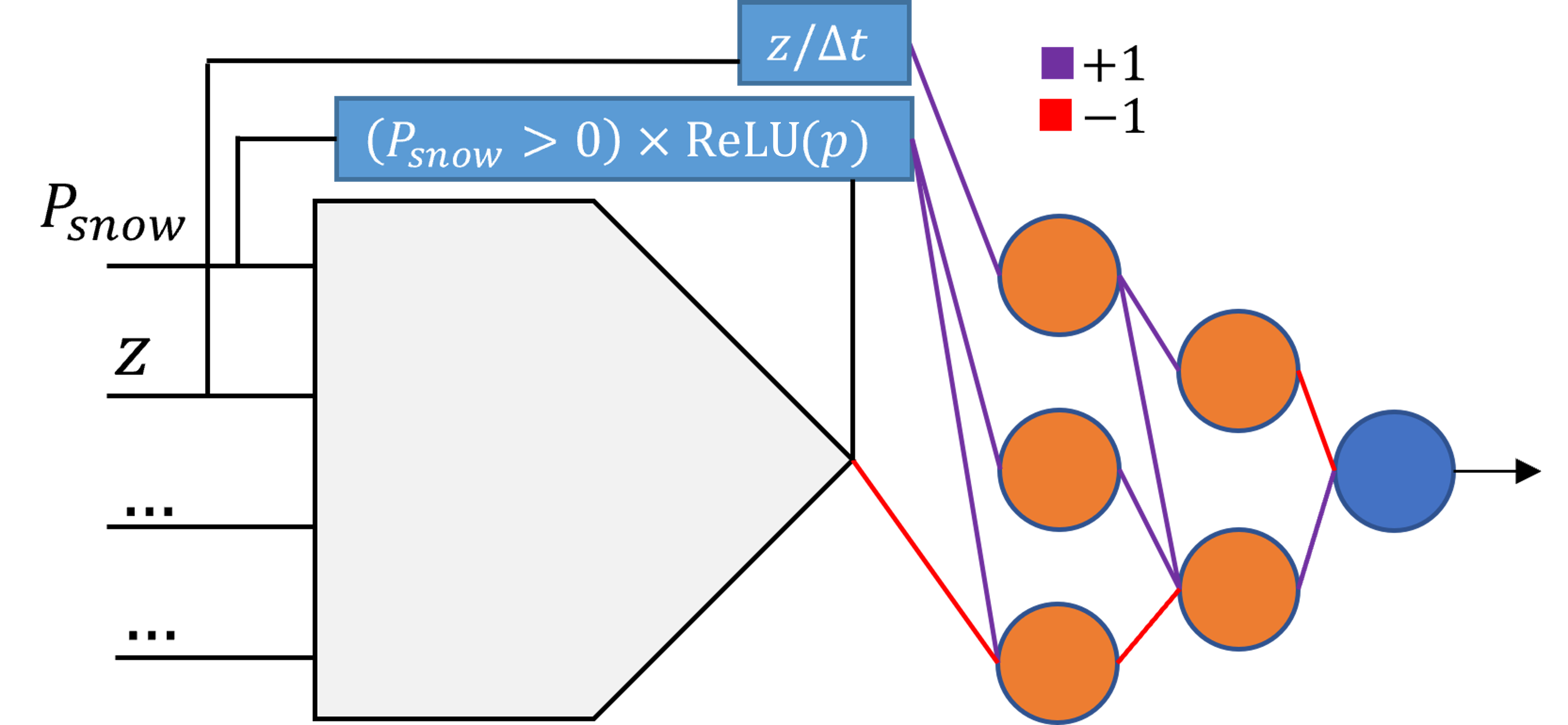}}
        \caption{Architecture of $M$, highlighting the constraint component attached to the predictive component (the grey pentagon) described in Fig \ref{f:pred}. The chosen structure enforces increasing snow depth only under precipitation and non-negativity of snowpack height, and is equivalent to a max/min block on the output. Weight colors indicate the constant's sign and activation functions follow the color scheme given in Fig.\ \ref{f:pred}.}
        \label{f:snowstruct}
\end{figure}

Constraints for $dz/dt$ should keep the depth tendency within physical limits to enable generalizability and stability when $M$ is integrated over time. This initial study selected the following basic constraints:
\begin{itemize}
\item Enforce depth non-negativity within a time step $\Delta t$, i.e., $M \geq -z/\Delta t$;
\item Enforce depth inability to increase without snowfall, i.e., $P_{\mathrm{snow}} = 0 \implies M \leq 0$. Processes like wind drift violate this constraint, but such effects are small in our data (see Appendix B).

\end{itemize}
These constraints can be expressed as threshold functions, the lower as $f_- = -z/\Delta t$ and the upper as $f_+ = {\rm{ReLU}}(p)\times\mathbf{1}_{\{P_{\mathrm{snow}}>0\}}$, where $p$ is the output of the predictive component and $\mathbf{1}$ is the indicator function. For these choices, $f_-$ is nonpositive and $f_+$ is nonnegative, which simplifies the constraint layer structure (see Appendix A), resulting in a final structure for $M$ as depicted in Fig.\ \ref{f:snowstruct}.

The first constraint includes the time step $\Delta t$, but this does not explicitly affect the time dependency or resolution of the parameterization. The predictive component contains no time nor time-step dependence. Adjusting $\Delta t$ scales the constraint appropriately without altering the predictive component’s output, enabling the model’s use in adaptive time-step schemes (re-scaling the constraint per time step). This means the model in principle only requires training with data at one temporal resolution, though we would anticipate improved time-step independence if training data incorporated varied time intervals, or $dz/dt$ values in the minimum range anticipated during usage.

\subsection{Data}\label{s:data}

\begin{figure*}[ht!]
        
        \centerline{\includegraphics[width = 39pc]{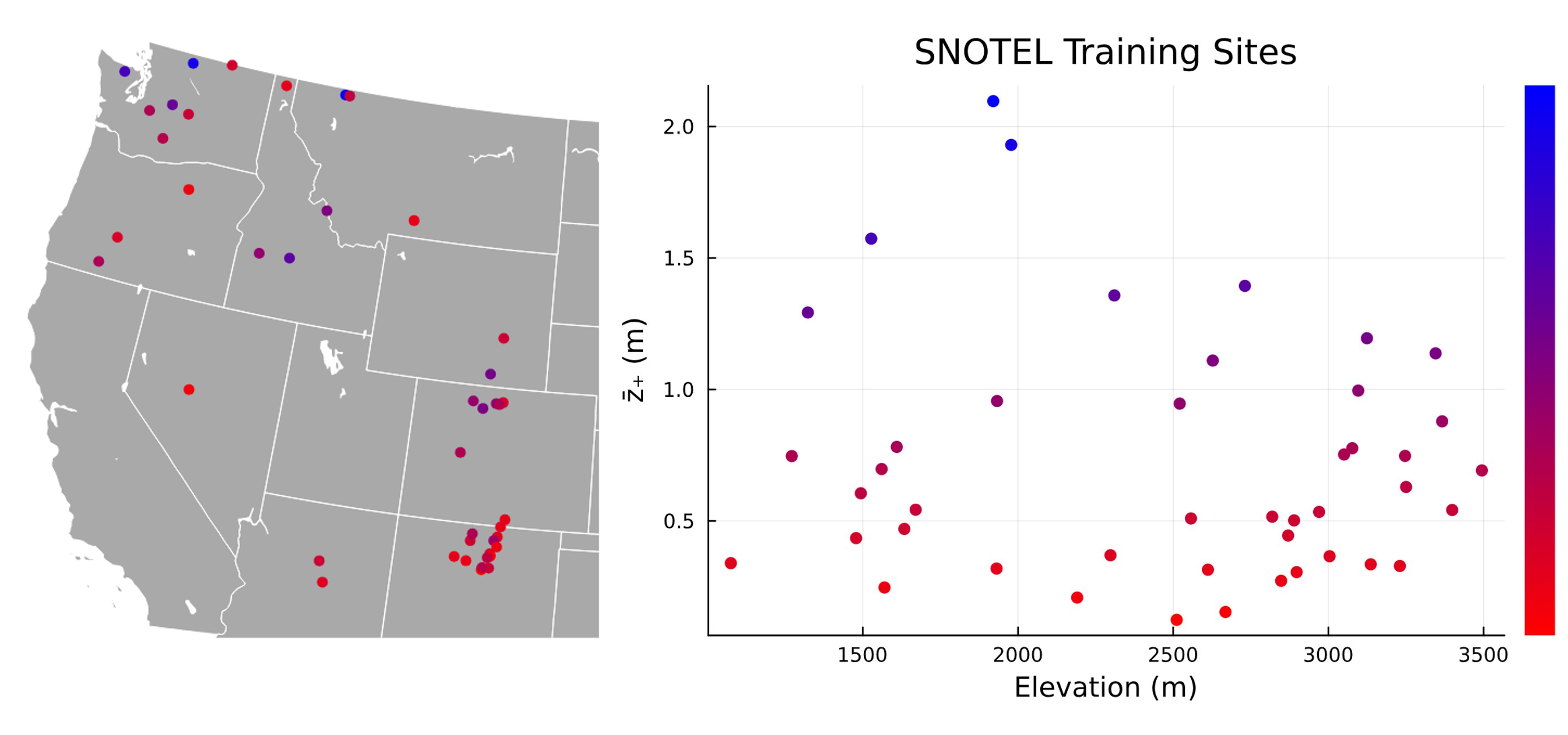}}
        \caption{Distribution of SNOTEL sites used for training the network. (a) Training sites as visualized over the United States. (b) Training sites visualized with elevation vs their average nonzero snowpack height, $\Bar{z}_+$. The color bar is a visual indicator of $\Bar{z}_+$ for visualization on the spatial map.}
        \label{f:locs}

\end{figure*}

We required training data with simultaneous snow and meteorological measurements, preferring collocated primary ground observations over reanalysis data due to known discrepancies \citep{joachin}. One primary source is the US Snow Telemetry Network (SNOTEL) by the Natural Resources Conservation Service (NRCS). Data from 44 total SNOTEL sites in the contiguous United States (CONUS) have simultaneous availability of the necessary inputs, of which the entire reporting history until February 1st, 2024 was collected. These sites span diverse climates (see Fig \ref{f:locs}), enhancing generalizability. For testing, 7 total Alaskan SNOTEL sites had the necessary inputs and were similarly collected, plus standard evaluation data from K{\"u}htai, Austria \citep{khutai_data}, Col de Porte, France \citep{coldeportdata}, and the Reynolds Mountain East catchment \citep{rme_data} to assess the $M$’s ability to generalize to out-of-sample data. We additionally used data from Sodankylä, Finland \citep{sodankyla}, the upper Rofental \citep{rofental}, and Yala Basecamp \citep{yala1, yala2} in the Himalayas to test performance across uncalibrated elevations and climate types.

SNOTEL data have known quality issues, such as underrepresenting complex mountainous terrain and underreporting precipitation \citep{meyer2012, serreze99}, along with periods of biased or unphysical values \citep{Hill19}. To ensure data suitability for model training, we applied established cleaning measures by \cite{serreze99} and refined by \cite{yan2018} to SNOTEL daily snowpack data. Gauge undercatch was corrected as in \cite{livneh} and temperature biases addressed following the SNOTEL correction release \citep{snotel-temp}. Since no consistent quality control exists for SNOTEL meteorological or snow depth data beyond outlier tests by \cite{Hill19}, we developed a custom procedure (see Appendix B). Snow fraction (to obtain snowfall from total precipitation) was estimated using the $(T_\mathrm{air}, \varphi)$  bivariate logistic model from \citet{rain-snow-split}, shown to have over 88\% accuracy. From the cleaned data, we derived $dz/dt$, $d\mathrm{SWE}/dt$, and $P_\mathrm{snow}$ for days with complete data, excluding all data with $\Delta t >$ 1 day.

The training data were averaged (preserving start-of-window $z$ and SWE) over a consecutive $N$-day moving window, with $N$ as a hyperparameter. This enabled exploration of the tradeoff of spreading out discretized SNOTEL data ($z$ to the inch, $\mathrm{SWE}$ to 0.1 inch) for smoother regression learning against preserving extreme values critical for predictions. Days with unphysical values (zero $\mathrm{SWE}$ and nonzero $z$) or no snowpack were removed to eliminate uninformative zeros in the target space. Features were scaled by their standard deviations, and the target by its absolute maximum, with scaling constants fixed in $M$ to spare user preprocessing. This resulted in 58,484 usable sensor-days out of 105,636 for training, and 35,618 sensor-days for testing. The complete dataset and generating code are publicly available (see Data Availability). 

\subsection{Training}\label{s:training}

Accurate prediction of extreme values is vital in snowpack modeling. Underpredicting extreme $dz/dt$ can prevent rapid snowpack growth or depletion, lagging snow presence early in the season or maintaining snow into the summer, which skews albedo, runoff, and energy calculations. Models like Noah, CROCUS, and SNOWPACK have struggled with this challenge \citep{china-late-pred, crocus2018-err, snowpack-lag, snowpack-lag2, crocus-lag}. Standard regression tends to under-predict extremes, so we used a custom loss function that can emphasize extreme values: 
\begin{equation} 
    L = \frac{1}{N_d} \sum_{i=1}^{N_d}  w_i \lvert y_i-\hat{y}_i \rvert^{n_1}\;,
    \;\;\;\; w_i = 1 + \lvert y_i \rvert ^{n_2}.
\end{equation} 
Here, $N_d$ is the number of batched training data, $\hat{y}_i$ and $y_i$ are the prediction and target data, and $n_1$ and $n_2$ are constant positive numbers. Using $(n_1 = 1$, $n_2 = 0$) or ($n_1 = 2$, $n_2 = 0$) is equivalent to optimizing the average $L_1$ or $L_2$ losses, respectively. Hyperparameter tuning followed a leave-one-out approach, using averaged and filtered data from 43 of 44 sites. Validation scores were generated over the remaining sites using unaveraged, unfiltered data, and averaged to guide hyperparameter selection. For timestepping (see section~2\ref{s:testing}), the optimal $n_1$ was found with $L_2$ training and $n_2 > 0$, highlighting the importance of extreme points. We note that optimal hyperparameters varied between regression and timeseries tasks; see Appendix C for more details. 

The model was implemented in Julia and the Flux framework \citep{Flux.jl-2018, innes:2018}, with the RMSProp optimizer \citep{RMSprop}. Training for 100 epochs (100 passes over all data) takes under 30 seconds on one Intel i9 CPU without GPU usage, with model storage under 3 kilobytes. Time and memory benchmarking of the network are listed in Table \ref{t:benchmarking} in Appendix C.

\subsection{Testing}\label{s:testing}
Model performance was tested by timestepping the $dz/dt$ equation with an explicit Euler method: 
\begin{equation}\label{eq:node-modified} 
    \hat{z}_{i+K} = \hat{z}_{i} + (K\Delta t) M(\hat{z}_i, \mathrm{SWE}_i, \varphi_i, R_i, v_i, T_{\mathrm{air},i}, P_{\mathrm{snow},i}). 
\end{equation} 
The integer $K$ specifies sequential data transitions ($K = 1$) or “gaps” to traverse in timeseries data ($K > 1$) due to missing or cleaned data. $M$'s built-in constraints ensure non-negative $z$ values when the step size is the designated $\Delta t$ (i.e. when $K = 1$; for providing other $\Delta t$ see Section 3f), though this choice of time-stepping procedure can create negative $z$ when $K > 1$ (rates constrained for step-size $\Delta t$ are applied over $K\Delta t > \Delta t$). In such cases, negative $z$ values were set to zero, and similarly, timeseries were “reset” to observed values  $\hat{z}_{i+K} = z_{i+K}$ whenever $K > K_\mathrm{max} = 5$ days, to avoid attributing error from the method choice to $M$ for fair evaluation. When $M$ is used within a global model simulation, such gaps would not occur since the inputs would be available at every step, allowing all $z_i$ to obey prescribed bounds.  Selecting $K_\mathrm{max} > 1$ day additionally reduced the number of resets that would otherwise beneficially skew performance metrics, with $K_\mathrm{max} = 5$ ensuring resets in less than 2.3\% of cases (median frequency 0.19\%), with many over whole years or no-snow periods (not impacting snow simulation). Evaluation metrics included root mean square error (RMSE), mean absolute error (MAE), bias (B), and median percent error (MPE, on $dz/dt$ for regression and $z$ for timeseries).  For timeseries, we included the Nash-Sutcliffe efficiency (NSE; \citealt{nash-sutcliffe}) and snowpack percent error (SPE), $\mathrm{MAE}/\Bar{z}+$, where $\Bar{z}+$ is the mean nonzero height.

By swapping SWE and $z$ features, the network can also be trained to predict $d\mathrm{SWE}/dt$ from $z$, allowing SWE to be simulated from depth data. This enables standalone modeling using only weather inputs, with two networks $\tilde{M}_z$ and $\tilde{M}_\mathrm{SWE}$, separately trained to run in a coupled fashion:
\begin{multline}
    \widehat{\mathrm{SWE}}_{i+K} =\\ \widehat{\mathrm{SWE}}_{i} + (K\Delta t) \tilde{M}_\mathrm{SWE}(\hat{z}_i, \widehat{\mathrm{SWE}_i}, \varphi_i, R_i, v_i, T_{\mathrm{air},i}, P_{\mathrm{snow},i}),
\end{multline}
and
\begin{equation}
    \hat{z}_{i+K} = \hat{z}_{i} + (K\Delta t) \tilde{M}_z(\hat{z}_i, \widehat{\mathrm{SWE}_i}, \varphi_i, R_i, v_i, T_{\mathrm{air},i}, P_{\mathrm{snow},i}).
\end{equation}
The only change to ensure physical consistency is to alter the lower bound of $\tilde{M}_z$ ($\tilde{M}_z$ = $M$ otherwise)  such that the $z$ update obeys $z_{i+K} \geq \mathrm{SWE}_{i+K}$ to enforce $z \geq \mathrm{SWE}$, so $\mathrm{SWE}_{i+K}$ was calculated first before $dz/dt$. This permits comparison with other models without inputting observational $\mathrm{SWE}$. 

To compare the neural model to established parameterizations, the Snow17 temperature-index model \citep{Snow17} was implemented and evaluated on the same data. Snow17 (designed for modeling runoff) models SWE and infers depth through Anderson’s density parameterization, and it can also assimilate observed SWE, allowing comprehensive comparisons. We performed two comparisons: (1) $M$ (the one-network depth parameterization) against Snow17 with both assimilating observational SWE, and (2) the two-network standalone model $\tilde{M}$ (subcomponents $\tilde{M}_z$ and $\tilde{M}_{\mathrm{SWE}}$) against Snow17, both using only meteorological inputs. To avoid confusion, Snow17 predicting versus assimilating SWE are labeled as ``SN17” and ``SN17O’, respectively. We used the Snow17 parameters from the \citet{wang2022} \textit{SN17-B-CONUS} model, calibrated over CONUS to account for regional climates. Both models were restarted with accurate depth when resets occurred, and computational benchmarking is compared in Table \ref{t:benchmarking}. Significance in differences between RMSE metrics were evaluated via a Wilcoxon signed rank test \citep{Wilcoxon} with $p=0.05$, which does not assume normal distributions or variance homogeneity.

In addition to depth, bulk density timeseries can be generated from $z$ and SWE data and model outputs according to Eq.\ \ref{eq:swe_def} and compared. The data is discrete while the model outputs are continuous, so densities were compared at sites with collocated $\mathrm{SWE}$ and $z$ sensors (see Appendix B) only when both models and data yielded physical values ($0 < \rho_{\mathrm{snow}}/\rho_{\mathrm{water}} < 1$). Counts of ``false non-snowpacks'' (models show $z=0$ while $z>0$ in the data) and ``false snowpacks'' (models show $z>0$ while $z=0$ in the data) were recorded, along with instances of unphysical densities.

To further assess physical consistency, the estimated rates $d\mathrm{SWE}/dt$ by $\tilde{M}_\mathrm{SWE}$ were compared to the snowfall rate data $P_{\mathrm{snow}}$. When the air temperature is below freezing ($T_{\mathrm{air}} < 0$), the conservation of mass implies these rates should be roughly equivalent, limited by factors such as runoff, sublimation, snow transport, and data precision. While it is possible to directly impose conservation of mass as a threshold in this framework, forgoing this specific bound allows for an investigation of the model's ability to represent physical constraints beyond those explicitly encoded.

Assessing physical consistency in the model directly per feature is nuanced due to strong correlations among the inputs, which confounds most interpretation methods like partial dependence, permutation, LIME, and tractable SHAP values \citep{interpml}. To isolate feature effects on model output, we calculated first-order Accumulated Local Effects (ALE) plots \citep{ALEplot}. This method shows the average change in model output accumulated over sequential bins of feature values. For feature $X$ at grid point $x_i$, data in the window $[x_i-\Delta, x_i + \Delta]$ are evaluated setting $X=x_i - \Delta$ and $X=x_i + \Delta$, storing the average of the differences $\Delta M _{x_i}= \overline{M(x_i+\Delta) - M(x_i-\Delta)}$. The final (centered) ALE value $\bar{\Delta M}$ at $x_i$ is the sum over all $\Delta M _{x_k}$ with $x_k\leq x_i$, minus the average of all uncentered ALE values. This isolates changes solely from feature variations, and bins are defined by quantiles to ensure equal data instances in each window. The shape and slope of the ALE curve are more pertinent for interpreting physicality than the offset, especially when the feature distribution is skewed. The range of $\Delta \bar{M}$ over the feature indirectly measures feature importance in influencing model output, as the ALE value can be interpreted as a departure from the average model prediction. Further information for interpreting ALE plots can be found in \citet{interpml}.

\section{Results}\label{s:Results}
\subsection{Depth Timeseries}
\begin{figure*}[ht!]
        
        \centerline{\includegraphics[width = 27pc]{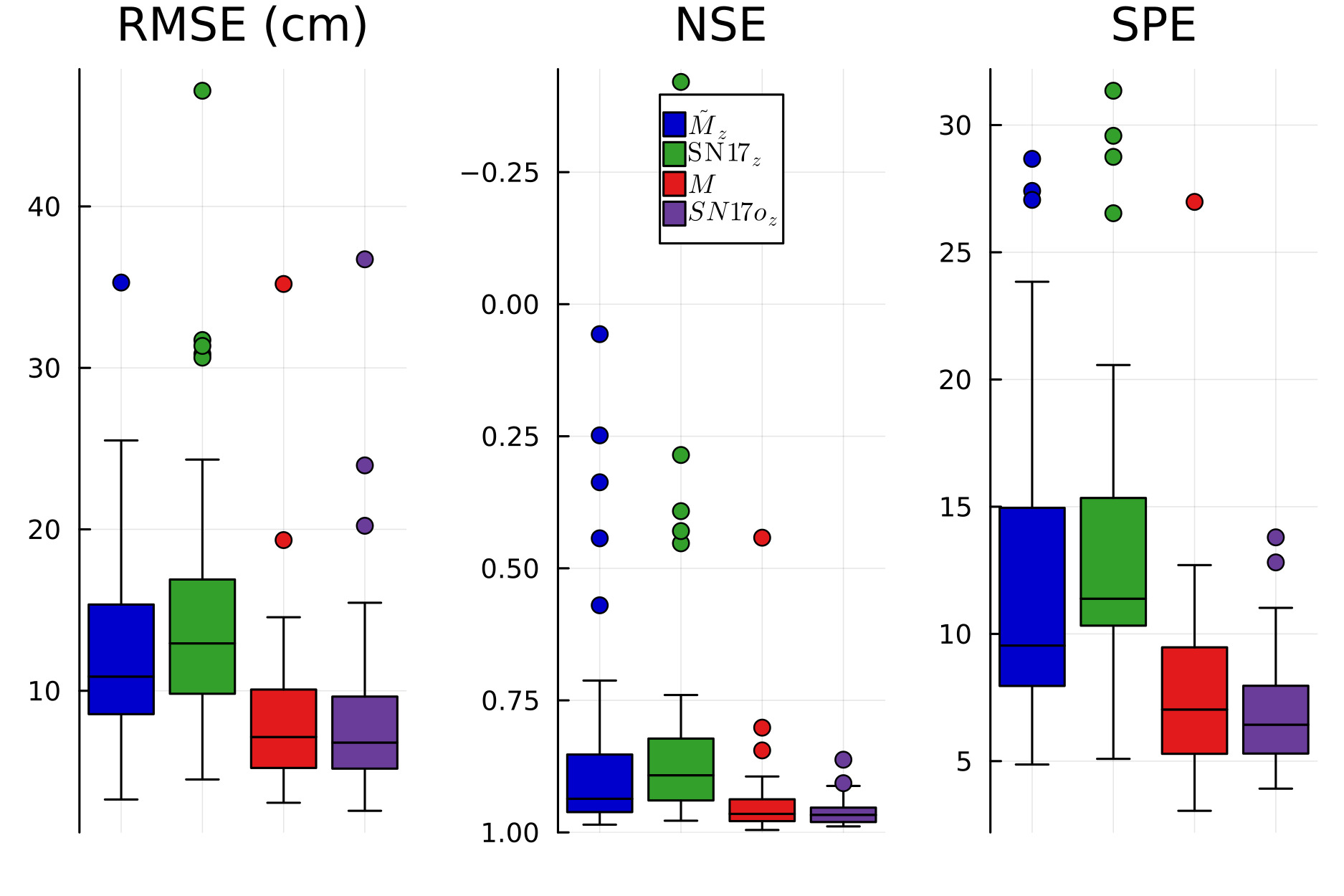}}
        \caption{Performance of $M$ and $\tilde{M}$ against Snow17 with (SN17O) and without (SN17) observational $\mathrm{SWE}$ data for generating $z$ timeseries over the 44 validation sites. RMSE indicates root mean square error, NSE the Nash-Sutcliffe Efficiency, and SPE the average L1 error normalized to the average nonzero depth, to measure percent error. Boxes outline the 25\% to 75\% quantiles, with the bar at the median, while whiskers mark the extremes and dots indicate outliers, which lie beyond 1.5 times the interquartile range (box width) from the box. Vertical axes limits are chosen to show all data points.}
        \label{f:validbox}

\end{figure*}

Over validation sites, neural configurations performed similarly to Snow17 independent of including observed SWE (Fig.\ \ref{f:validbox}). Both models exhibit similarly tight spreads, though $M$ exhibited larger spreads than SN17O when utilizing SWE data. Conversely, when simulating SWE, the spread was larger for SN17 compared to the coupled neural model $\tilde{M}$, which is more indicative of usage under SWE uncertainty or provision of SWE within a separate model. Statistically significant improvements in RMSE were found for both SWE and $z$ modeling by $\tilde{M}$ compared to SN17 ($p = 0.029$ and $p = 0.0006$, respectively). No significant difference was observed for $M$ against SN17O using SWE data for depth parameterization alone ($p = 0.673$), which is unsurprising as both models were calibrated for performance over the locations represented in this data.

\begin{figure*}[ht!]
        
        \centerline{\includegraphics[width = 27pc]{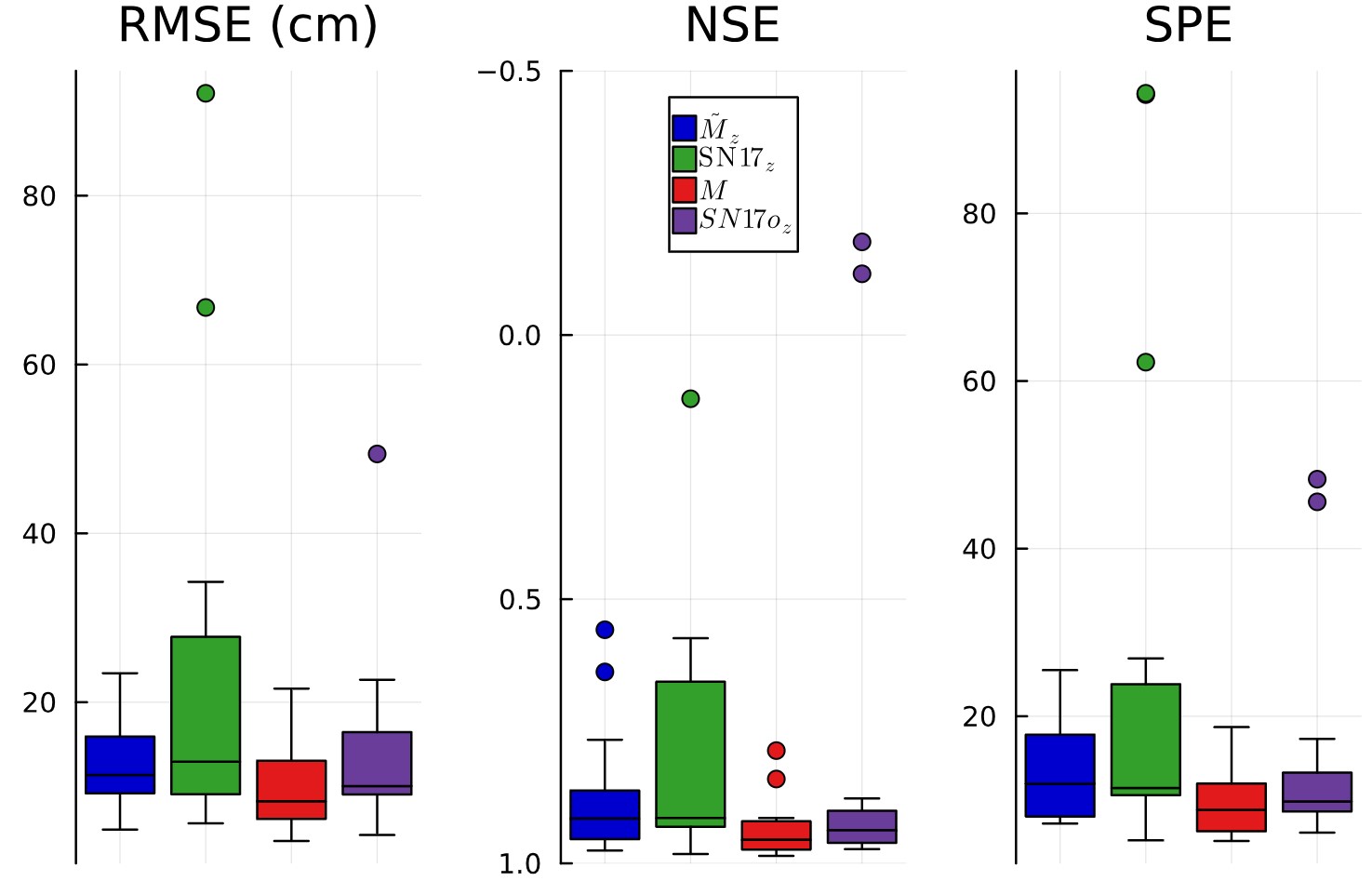}}
        \caption{Performance of the neural parameterization against Snow17 over the 14 testing sites in depth timeseries generation. Labeling convention follows that in Fig.\ \ref{f:validbox}. One outlier for SN17 with an NSE of -2.8 is not shown on the plot to aid in the scaling of the other values.}
        \label{f:testbox}

\end{figure*}

\begin{figure*}[ht!]
        
        \centerline{\includegraphics[width = 39pc]{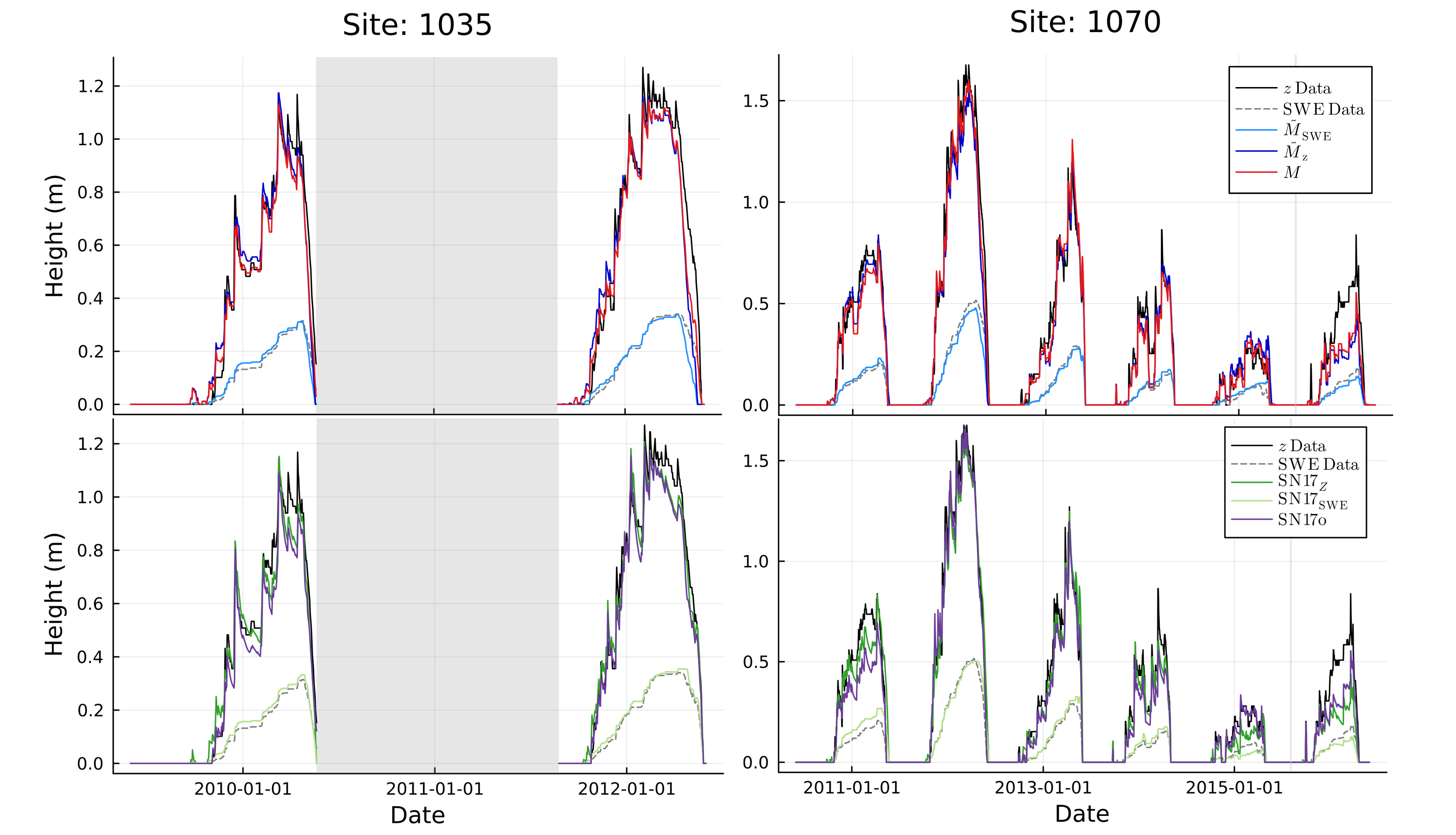}}
        \caption{$z$ and SWE timeseries over two subsets of the Alaskan testing sites. Data gaps (missing or cleaned data) are shaded in grey. All models perform similarly, though the neural models seem to perform poorly whenever Snow17 also performs poorly, like the winter of 2016 at Site 1070. This suggests the neural parameterization is at least an efficient surrogate for Snow17, or could insinuate a data inconsistency if both models fail similarly.}
        \label{f:alaska}

\end{figure*}

The model performance over all testing sites is summarized in Fig.\ \ref{f:testbox}, with example timeseries in Fig.\ \ref{f:alaska}, and medians and averages reported in Appendix C (Table \ref{t:performance}). Although the performance distributions are non-symmetric, both mean and median metrics are important for gauging potential (median) and consistency (mean) for generalizing to out-of-sample regimes. The neural configurations showed tighter performance spreads across testing sites and climates not included in the training data, despite no recalibration. Average neural NSE was 0.87 and 0.94 (without and with SWE data, respectively), compared to 0.35 and 0.78 for Snow17 --- median differences were smaller, but still favored neural models. Snow17’s mean SPE was 15\% even with observational SWE, comparable to other established models without site-specific calibration \citep{crocus2012-err, crocus2013-err, Crocus2020-err, crocus2018-err, ski-resort-err, nunavut_ML_err, china-late-pred, bulk-density_models}, and nearly doubled to 28\% without SWE data. In contrast, the neural parameterization maintained mean SPE under 15\%, increasing modestly from 9.5\% to 13.4\% without SWE data, demonstrating greater robustness and consistency out-of-sample despite the simplicity of its predictive component and lack of past snow depth state storage. 

Model $M$ demonstrates strong generalizability, with similar performance and tighter spreads across test and validation sites, without retuning. Although the neural model did not exhibit significant RMSE improvements over Snow17 on out-of-sample data at the $p=0.05$ level ($p = 0.14$ for $z$, $p = 0.27$ for SWE), it did outperform Snow17 as a parameterization ($p = 0.01$). This suggests that when paired with a reliable SWE predictor within a larger model, the neural approach offers advantages over prevailing alternatives in extrapolating to uncalibrated locations.
\subsection{Density Timeseries}
\begin{table*}[ht!]
\caption{Results of bulk density timeseries generation for each model. The median score over all validation and viable testing sites is presented, and scores are derived from all predictions of physical densities during observed snowpacks, or otherwise tallied in the presented counts. The integer count gives the median number of occurrences across sites, while the percentage normalizes each count against the length of the timeseries.}
\label{t:dens}
\begin{center}
\begin{tabular}{ccccc}
\topline
Parameter & $\tilde{M}$ & $M$ & SN17 & SN17O\\ \midline
RMSE (\%)& 9.1 & 9.5 & 7.4 & 7.6\\
Bias (\%)& -1.9 & -1.8 & 0.6 & 0.9 \\
MPE (\%) & 11.3 & 12.9 & 14.1 & 14.9\\
False Non-Snowpacks & 4 (0.13\%) & 2 (0.08\%) & 14 (0.68\%) & 0 (0\%)\\
False Snowpacks & 183 (8.6\%) & 136 (6.1\%) & 45 (3.2\%) & 2 (0.10\%)\\
Unphysical Points & 3 (0.17\%) & 3 (0.12\%) & 0 (0\%) & 0 (0\%)\\
\botline
\end{tabular}
\end{center}
\end{table*}

\begin{figure*}[ht!]
        
        \centerline{\includegraphics[width = 39pc]{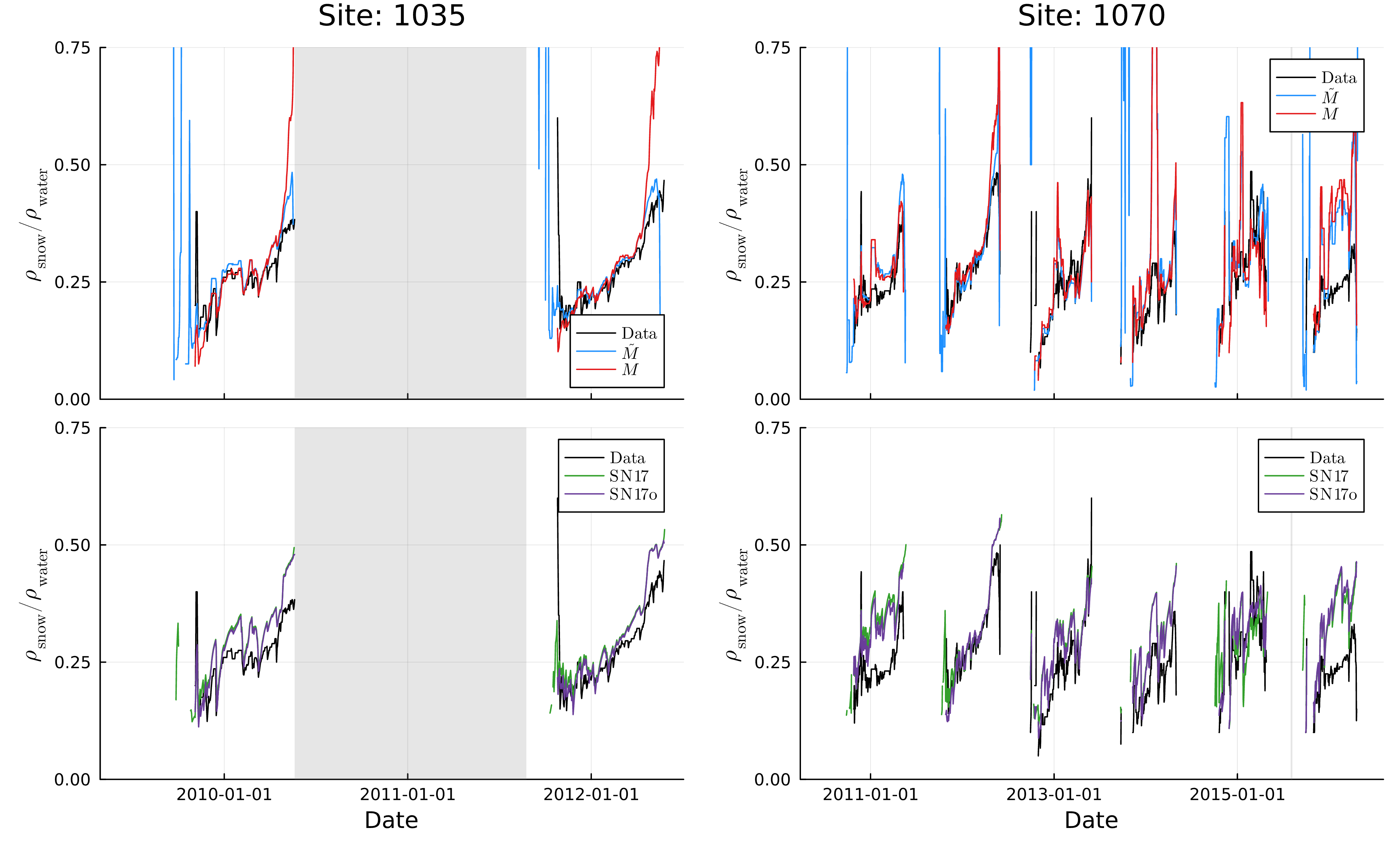}}
        \caption{Bulk density timeseries (normalized against $\rho_{\mathrm{water}}$) over the same series as Fig.\ \ref{f:alaska}. Poor representation at the start/end of the season for small snowpacks are exacerbated by data discretization and lack thereof in the models, particularly in $\tilde{M}$ and SN17, or from simpler constraints in $M$ and $\tilde{M}$. Such fluctuations severely skew the NSE and SPE metrics and confound their measure of model performance. Timeseries are only shown where models and data show a relative density between 0 and 1.}
        \label{f:density}

\end{figure*}

For validation and testing sites with collocated $z$ and SWE measurements (see Appendix B), Table \ref{t:dens} compares timeseries statistics for derived bulk density, and Fig.\ \ref{f:density} displays estimates from the same data as Fig.\ \ref{f:alaska}. RMSE values reflect relative error since density is normalized by $\rho_{\mathrm{water}}$. Both SN17 and $\tilde{M}$ lack SWE discretization, inflating errors for small snowpacks compared to their assimilated counterparts. Neural unphysical densities source from a lack of constraint enforcing $\mathrm{SWE} = 0 \implies z= 0$ (in both $\tilde{M}, M$) or $z > \mathrm{SWE}$ (in $M$), though different constraint choices could eliminate this. Snow17 is roughly $3\times$ better at mitigating false snowpacks with lower RMSE (less extreme errors), but our models are roughly $3\times$ better at reducing false snow-absence and exhibit improved individual density predictions, crucial for estimating other snowpack properties. Thus, the advantage between models depends on which features are prioritized. Otherwise, the two perform similarly, despite $M$ being designed for snow depth while the comparable parameterization within Snow17 is explicitly formulated for density. Among evaluated sites, the best and worst performances by $M$ were better than those of SN17O. 
\subsection{Predicted $dz/dt$}
\begin{figure*}[ht!]
        \centerline{\includegraphics[width = 39pc]{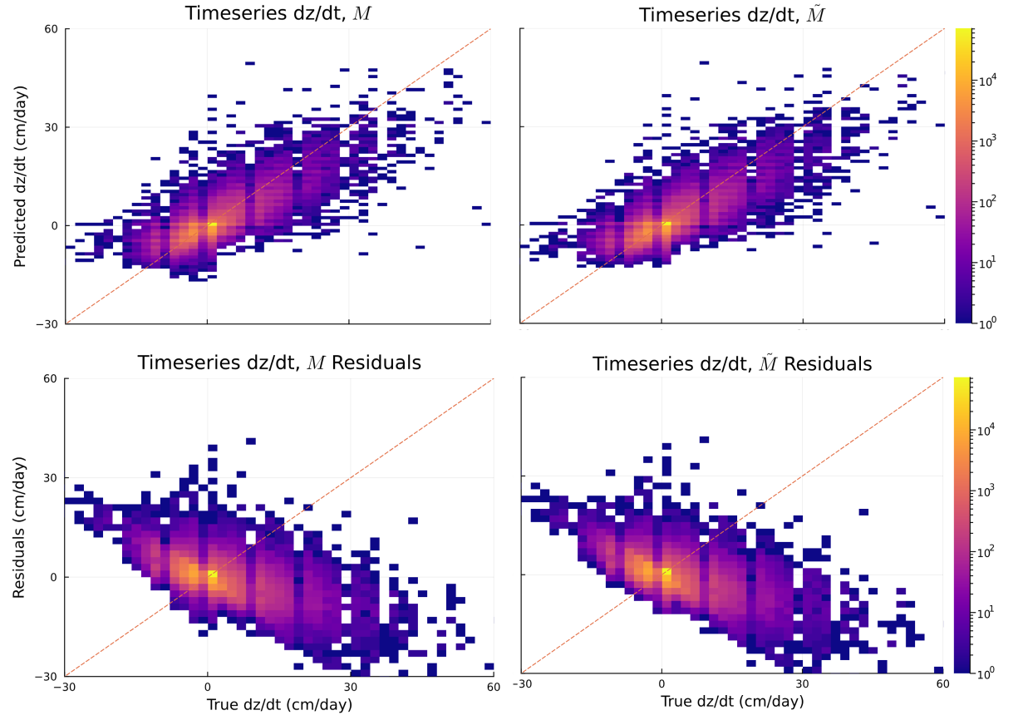}}
        \caption{Predicted vs.\ true $dz/dt$ and residuals against the modeled target by the $M$ and $\tilde{M}$ models. Both models continue to under-predict extremes, despite the bias given to extremes during training. The magnitude of data in small $dz/dt$ ranges relative to extremes could contribute to this phenomenon.}
        \label{f:scatter}

\end{figure*}

Fig.\ \ref{f:scatter} shows a histogram comparing predicted versus true $dz/dt$ values from $M$ (Pearson correlation $r = 0.78$) and $\tilde{M}$ ($r = 0.77$) during timeseries generation. Both still display a tendency to under-predict extreme values. This is likely from exposure to much more training data with small $dz/dt$, which could result in better predictions of small values at the expense of extremes without applying other methods like class balancing. 

\subsection{Physical Behavior of Model}
For SNOTEL instances with $T_{\mathrm{air},i} < 0$, $\tilde{M}_{\mathrm{SWE}}$ violated the $d\mathrm{SWE}/dt \leq P_{\mathrm{snow}}$ condition only 0.6\% of the time, with all violations under 1 mm/day and 95\% under 0.75 mm/day. About half the instances showed $d\mathrm{SWE}/dt < P_{\mathrm{snow}}$ (prescribed bounds yielded equivalence in the rest), but the mean residual of these was -2 mm/day, with 95\% above -5 mm/day, aligning with typical sublimation rates \citep{sublimate1, sublimate2} and effects not explicitly modeled. Given the SNOTEL data precision (2.54 mm) and negative average temperatures obscuring instantaneous positive temperatures, these results suggest strong adherence to mass conservation, even without explicitly enforcing it (as simple as choosing $f_+ = P_{\mathrm{snow}}$ from Section 2\ref{s:constraints} for $\tilde{M}_\mathrm{SWE}$), highlighting the model's ability to learn physical representations from data beyond those prescribed.

\begin{figure*}[ht!]
        \centerline{\includegraphics[width = 39pc]{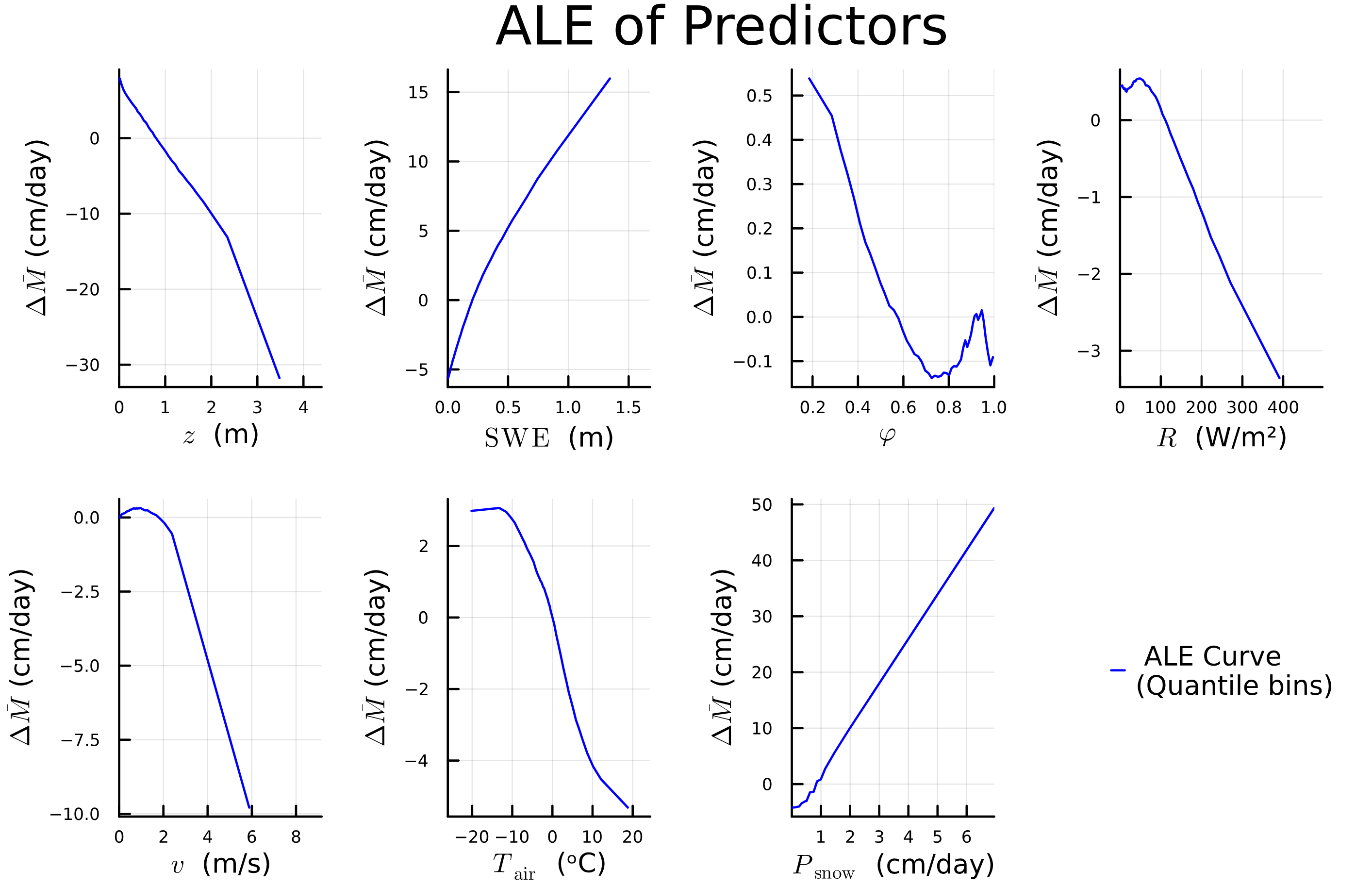}}
        \caption{Accumulated Local Effects (ALE) plot for $M$'s predictive features. The ALE mitigates effects of feature correlations, providing a rough indication of feature importance in determining $M$'s output magnitude and a visual measure of physicality. x-axes mark the range of each feature, while y-axes show the change in $M$ relative to the average prediction. Curves are binned by quantiles so each bin has at least 50 samples.}
        \label{f:ale}

\end{figure*}

The ALE plots for each feature are shown in Fig.\ \ref{f:ale}. The average prediction of $M$ is $\approx 0$, with negligible centering offsets, indicating a tendency to predict negative $dz/dt$ after $T_{\mathrm{air}} > 0$. $M$ also exhibits a linear relationship in $dz/dt$ with $P_{\mathrm{snow}}$ and stabilization at low $ T_{\mathrm{air}}$ (where a minimum snowfall density would emerge), all aligning with physical expectations. Additionally, decreases in $dz/dt$ are observed with increased solar radiation and wind speeds, reflecting the model's learning of understood destructive processes. However, data availability and artifacts may influence these results; for instance, while higher relative humidity could lead to more condensation and surface melt, it also lowers estimated snowfall fraction under the data preparation used. Moreover, snow accumulation on sensors during heavy snowfall may produce saturated humidity readings linked to large $dz/dt$, and location biases may affect the value ranges. 

\begin{figure*}[ht!]
        
        \centerline{\includegraphics[width = 39pc]{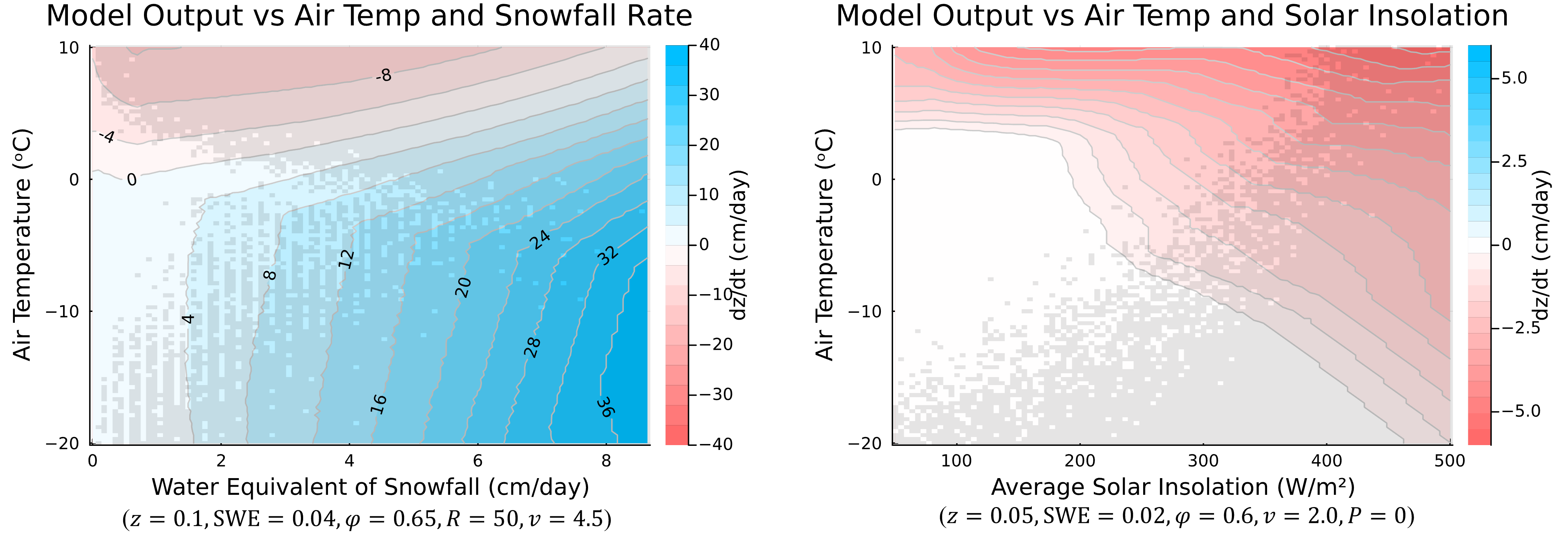}}
        \caption{$M$ outputs for two snow states, with other inputs held constant. In each case one threshold ($dz>-z$ or $dz/dt < 0$ for $P_\mathrm{snow}=0$) is visible. ``Shaded'' or ``masked'' pixels (partially greyed out) indicate areas of the visible parameter space with ranges not represented in the data, either from being unphysical or unobserved.}
        \label{f:derivatives}

\end{figure*}

$M$ can be directly queried over ``slices" of its multidimensional input space, facilitating exploration of the model's behavior in anticipated input regions that may not be observed in the current data. Unlike partial dependence plots, these slices are not averaged over training data ranges, allowing them to display outputs in regimes of inputs that are not physically viable. Additionally, patterns observed in one ``slice” may not be conserved across others. Fig.\ \ref{f:derivatives} illustrates these slices, with areas lacking data slightly masked.

In Fig.\ \ref{f:derivatives}a, contours reflect physical expectations, with even spacing indicating $dz/dt$ is linear in $P_{\mathrm{snow}}$, $T_\mathrm{air}$ becoming the dominant feature around $T_\mathrm{air}=0$, and the constraint $\Delta z \geq -z$. Fig.\ \ref{f:derivatives}b examines how $T_\mathrm{air}$ and insolation impact snow depth at zero snowfall, showing depletion begins once $T_\mathrm{air} > 0$ and with increasing $R$. All output values at zero snowfall remain nonpositive, adhering to the prescribed threshold. This slice demonstrates limited sensitivity to solar insolation at low $R$, suggesting $T_\mathrm{air}$ is the dominant variable. This insensitivity may arise from high snow reflectivity at low incidence angles (and normal sensor orientation), shading effects reducing melting until higher radiation levels are exposed, or latent melting effects prevailing at low irradiance. Positive feedback loops where accumulating surface melt alters albedo could explain the transition in this slice. Overall, the model aligns with expected physical behavior over available data, though incorporating more data into extrapolated (shaded) regions could enhance universality.

\subsection{Generalizability}
\begin{figure*}[ht!]
        \centerline{\includegraphics[width = 39pc]{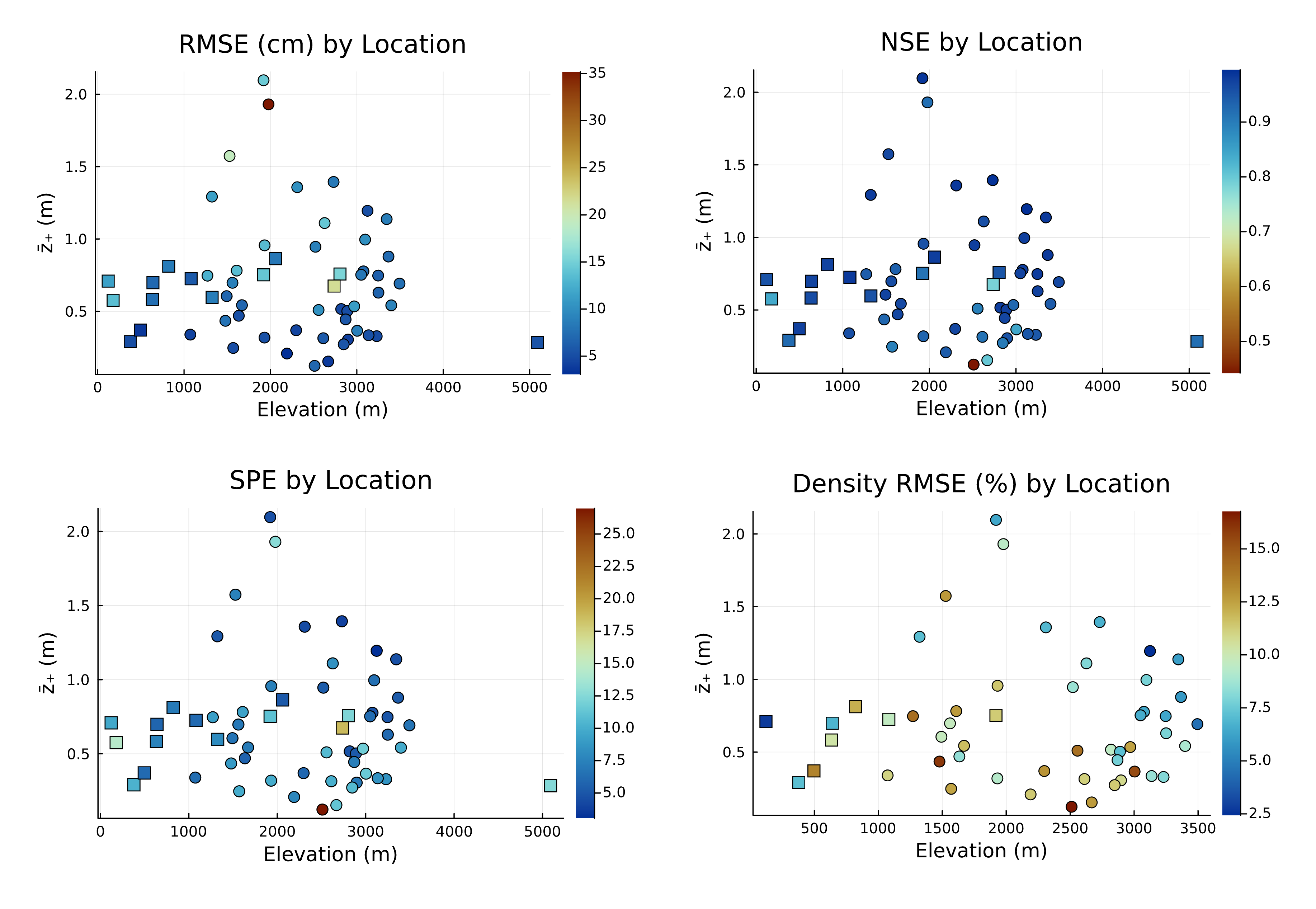}}
        \caption{Performance of $M$ across all training and testing sites with regards to RMSE, NSE, and SPE of depth timeseries, as well as RMSE for density timeseries over the sites where density was evaluated (see Appendix B). Testing sites are marked with squares instead of circles.}
        \label{f:perf}
        
\end{figure*}

To assess generalizability, Fig.\ \ref{f:perf} shows elevation vs.\ mean nonzero snow depth $\Bar{z}_+$ for all training and testing sites, similar to Fig.\ \ref{f:locs}. Sites are colored by the performance of $M$ for SPE, RMSE, and NSE on $z$ timeseries, along with density RMSE. The model succeeds comprehensively, with SPE errors under 20\% for nearly all sites and most under 10\%, while density errors remain predominantly below 15\% RMSE. It performs comparably on testing data from different elevations and climates, indicating robust generalizability, even for density calculations. No discernible trend with elevation appears, corroborating generalizability rather than elevation-induced effects. 

\subsection{Finer Resolution Predictions}
\begin{figure}[ht!]
        \centerline{\includegraphics[width = 19pc]{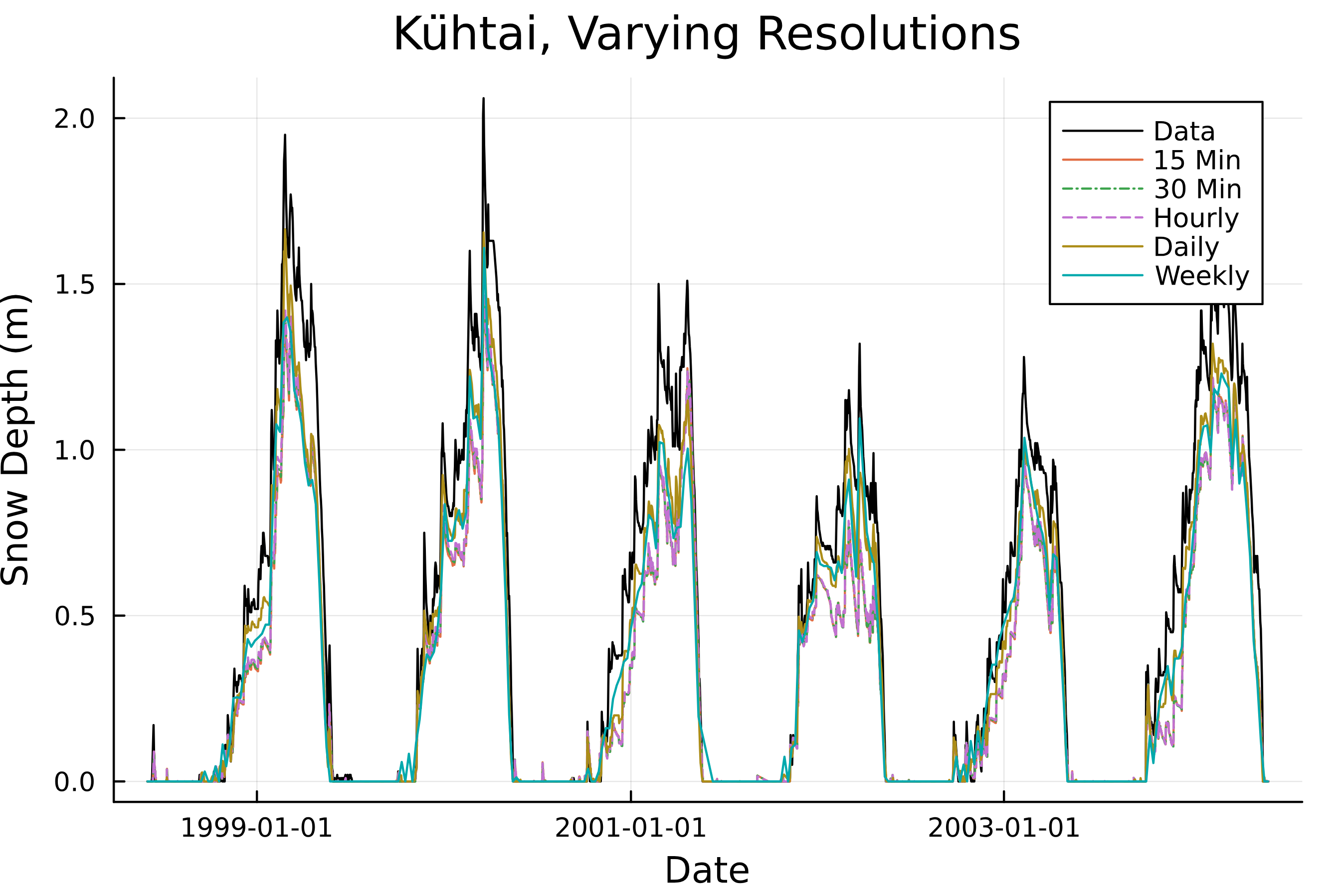}}
        \caption{Output of $M$ at different resolutions for a subset of the site data from K{\"u}htai, Austria. The graph overlays outputs at different resolutions for direct comparison. Sub-hourly (15 min, 30 min) curves are nearly indistinguishable from the hourly curve.}
        \label{f:resolution}
\end{figure}

Time units only appear as rates in $M$ via precipitation, the lower bound, and the output. By predicting rates $dz/dt$ instead of accumulated $dz$, $M$ can be evaluated at varying time-steps without retraining, by merely resetting the constraint function's scaling of $1/\Delta t$ without altering the trained predictive weights (which were trained only once in section 3a, with $\Delta t$ = 1 day). This flexibility allows testing at resolutions up to data limits, which for K{\"u}htai is a maximum resolution of 15-minute intervals. While sub-hourly time resolutions are rarer in larger earth models, they are used in land models and will become increasingly frequent as research advances in fine-scale land-atmosphere interactions \citep{schar, Ban2021}. Similarly, while multiday resolutions are rare, data availability can necessitate their use. Benchmarking across scales above and below the daily resolution used for training offers insights into the model's limits and potential applications. However, beyond a week, average input variables do not effectively capture critical input dynamics (e.g., monthly average $T_\mathrm{air}$ would not reliably indicate time above freezing), leading to suppressed variations that undermine meaningful outputs, and alternative model choices would be more appropriate. Results of repeating the timeseries generation with $M$ as described in Section 3f for the K{\"u}htai data with different values of $\Delta t$ in the constraint but identical weights otherwise, spanning 15-minute to weekly resolutions, are shown in Fig.\ \ref{f:resolution} and Table \ref{t:resolution}.

\begin{table*}[ht!]
\caption{Performance of $M$ at varying resolutions for $z$ timeseries generation. Error jumps beyond the daily training resolution, but performance remains near-constant between hourly and 15-minute resolution.}
\label{t:resolution}
\begin{center}
\begin{tabular}{cccccc}
\topline
Statistic/Resolution & Weekly & Daily & Hourly & 30-Minute & 15-Minute \\ \midline
RMSE (m) & 0.1762 & 0.1437 & 0.2008 & 0.1989 & 0.1999 \\
NSE & 0.759 & 0.806 & 0.708 & 0.710 & 0.709 \\
SPE (\%) & 23.90 & 19.54 & 30.9 & 31.0 & 31.1 \\ \botline
\end{tabular}
\end{center}
\end{table*}

The timeseries shown at this site exhibit a low bias; however, for all resolutions, $M$ achieved over a 40\% reduction in RMSE compared to Snow17 (not tabulated). RMSE increases at timesteps outside the daily interval used for training, though all sub-hourly resolutions yield nearly identical results without further trends. This performance loss may arise from the extreme values of $dz/dt$ and precipitation observed at higher resolutions. While daily data show gradual snowpack increases, finer resolutions might capture the same deposition over a few hours, leading to $dz/dt$ values 10–20 times larger than those in the training data. Conversely, weekly averages smooth out extreme events, which can reduce the variance of outputs. Overall, $M$ demonstrates an ability to transfer across temporal resolutions, though performance would likely improve with a broader range of $dz/dt$ training values (for instance, incorporating both hourly and daily data). 

\section{Discussion}\label{s:discussion}
This study explored a simple, versatile data-driven framework to enhance physical parameterizations, focusing on generalizable snow parameterizations for climate modeling applications. Many choices were results-driven and informed by data availability, such as selecting widely measured variables to increase model applicability, though other choices could likely enable representation of additional processes. Data requirements limited sources to the SNOTEL network, which, while useful for local relationships, has quality issues compared to validation sites such as Col de Porte and K{\"u}htai \citep{meyer2012, joachin}, fails to represent large-scale heterogeneity effects, especially in mountain regimes \citep{joachin}, and lacks coverage of extreme conditions in scarcely sampled tundra and taiga biomes. The approach's ability to extend to these unsampled terrains or perform on coarser grids with explicit large-scale effects within an Earth system model remains untested and will be the subject of a future paper. Further adaptation of these models in a world of growing data volume, frequency, and quality offers an exciting opportunity for future research. 

The neural model $M$ generalizes well across locations, a crucial feature for both global consistency and local-scale modeling, particularly as climate change skews site statistics to become ``new” locations. Snow17 struggles with such shifts, limiting its application on longer timescales \citep{joachin, boone}. Our physics/ML hybrid approach offers a viable alternative. Generalizability is key for ML in climate models. For instance, \cite{yang2020}'s random forest model applied to Chinese sites showed high out-of-sample biases and SPE compared to our model, despite similar RMSE. Their ML approach and others like \cite{duan} or \cite{yang2022} rely on location-specific variables (slope, aspect, topographic/vegetative indices, etc.) or features like historical averages and microwave measurements \citep{LSTM1, lstm2, lstm3, rf2, yang2022}, which hinder usage within earth system models. In contrast, \cite{wang2022} simulated SWE with recurrent neural networks using physical inputs with similarly high NSE scores and moderate generalizability, but required over 240 previous states for updates. \cite{duan} also used 180 days of forcing data per SWE prediction with various models at SNOTEL sites, requiring training times of 5–26 hours and hours of simulation time on a GPU. Our memoryless neural ODE model achieves comparable results to \citet{wang2022} with guaranteed consistency with physical bounds. It exhibits improved median MAE, RMSE, and NSE against all models from \cite{duan}, with fewer inputs and significantly fewer computational resources (see Table \ref{t:benchmarking}). \cite{Steele} used the same inputs as ours for a standalone ML model and a post-processing model for physical models. Both yielded higher SWE RMSEs (6 cm and 13 cm vs. $\sim$4 cm in our model, see Table \ref{t:performance}) and poorer generalization, though a slightly better derived density RMSE (implying $z$ errors were similarly scaled to SWE errors). Both models required addition of a binary snow-presence variable to reset unphysical summer snowpacks, while our approach naturally adheres to prescribed bounds, eliminating such unphysical departures.

The best hyperparameters for timeseries generation differed from those for direct regression, underscoring better $dz/dt$ predictions do not guarantee better accumulated seasonal timeseries. This further supports that our choices are going beyond merely matching magnitudes, and instead summarizing universal, memoryless physical processes. Notably, the parameterization faltered only when new climates introduced target magnitudes absent in the training data, rather than when locations presented different input feature magnitudes. While output magnitude extrapolation is limited as with many data-driven models, the input generalizability is a less common result, highlighting the benefits of enabling physical consistency. This underscores the need for more widespread snow sensing, particularly in extreme climates, to improve predictive power of such models. 

The framework’s application of prescribed constraints around black-box models like neural networks enables easy modification of constraints or input variables. This interchangeability supported rapid experimentation and prototyping, and is synergistic for integration as a ``plug-and-play" model that can adapt to available inputs and operational constraints. The approach provides linear scaling in input size with low computational overhead, which can reduce computational budgets while maintaining or improving accuracy compared to advanced models, permitting simulations longer into the future or over finer grids. It can demonstrably act as a standalone predictive tool wherever inputs can be measured or inferred, which could be through observations, remote sensing, weather forecasts, or coupled models. This framework could be used, for example, for forecasting applications such as weekly skiing or hiking terrain predictions from weather forecasts, or tested for water supply simulation given snowpack data. The model structure is a strong candidate for simulating many types of constrained physical systems beyond snow, offering further avenues for investigation (see Appendix A). 

Beyond testing the neural model $M$ globally in a coupled setting with an entire climate model, other future directions of research could involve adapting $M$ to continuous neural ODE structures \citep{neuralode}  or more general timestepping schemes. While this study focused on localized relationships and comparisons with similarly-formulated standards, it did not address large-scale heterogeneity effects. Future adaptations to directly incorporate these effects might aid usage within coarser-scale (${\sim}100$~km) models. Additional improvements could involve integrating data from NOHRSC/MADIS or detailed snow layer data to simulate temperature profiles. Alternative training strategies, such as using timeseries error as the loss function or gradient-free update rules to bypass recursive timeseries gradient issues could also be explored. 

\section{Conclusion}
Using a location-agnostic and physically constrained neural ODE framework to parameterize the rate of change of snow depth, we were able to simulate seasonal snow depth with a median error of 8.8\% across sites with varying climates and elevations, including some not seen during training. Though the parameterization was trained with daily data, it shows an ability to perform with moderate accuracy at other temporal resolutions without additional retraining of the model; however, retraining with higher-resolution training data may lead to further improvement. The parameterization's structure reduces computational overhead while maintaining performance in depth simulation at the level of established, cutting-edge, or more detailed models. The design is conducive for usage in prognostic models or can be adapted to alternatively predict variables such as $\mathrm{SWE}$. When driven solely by meteorological data as a standalone model, the parameterization framework can recreate seasonal timeseries with comparable error without retraining or site calibration—--an improvement over other established models. In most cases, it matches or outperforms the Anderson Snow17 model in simulating seasonal snow depth, offering an efficient formulation for use within physical models and an alternative to prevailing parameterizations.

The proposed framework demonstrates potential for a wide array of applications for both long-term climate simulations as well as short-term forecasting applications. The means of enforcing hard constraints structurally provides a simple but powerful technique for predictive modeling that can be applied beyond snowpack modeling to different climate processes or physical parameterizations.

%
%
%

%

\clearpage
\acknowledgments
We thank Marie Dumont for insightful discourse on process-based snow models, the SNOTEL effort, and the K{\"u}htai, Col De Porte, Reynolds Mountain East, Sodankyla, Rofental, and Yala Basecamp teams for their data. A. C. was supported by the AI4Science initiative at the California Institute for Technology and a Department of Defense National Defense Science and Engineering Graduate (NDSEG) Fellowship. This work was generously supported by Schmidt Sciences, LLC, and  the Resnick Sustainability Institute. The authors thank Jeffrey Coyle, Jaz Ammon, Joseph Kral, Matt Warbritton, and Daniel Tappa for help in verifying SNOTEL sensor placement, and Yuan-Heng Wang for sharing calibrated Snow17 parameters.


%
%
\datastatement
The SNOTEL data utilized in this study was available via the National Water and Climate Center, which lies under the United States Department of Agriculture. Data reports of the SNOTEL data were generated using the online portal found at \url{https://www.nrcs.usda.gov/wps/portal/wcc/home/}. The data from Col De Porte \citep{coldeportdata} can be found at the Observatoire des Sciences de l’Univers de Grenoble DOI portal \url{https://doi.osug.fr/public/CRYOBSCLIM_CDP/CRYOBSCLIM.CDP.2018.html}, and data from K{\"u}htai \citep{khutai_data} can be found as the supplementary material from \url{https://doi.org/10.1002/2017WR020445}. Raw data from Sodankyla \citep{sodankyla} can be found from the Intensive Observation (sensors 8, 11) data portal \url{https://litdb.fmi.fi/ioa.php} and automated weather station (sensor 15, portal \url{https://litdb.fmi.fi/luo0015_data.php}. Reynolds Mountain East \citep{rme_data} data was obtained from an ESM-SnowMIP repository \url{https://www.geos.ed.ac.uk/~ressery/ESM-SnowMIP.html}. Raw Yala Basecamp \citep{yala1, yala2} data was retrieved from \url{https://rds.icimod.org/Home/DataDetail?metadataId=26859} and \url{https://rds.icimod.org/Home/DataDetail?metadataId=1972554}. Rofental \citep{rofental} data was retrieved from \url{https://datapub.gfz-potsdam.de/download/10.5880.FIDGEO.2023.037-MNveB/}.

The data in this study was processed from these sources. Code for scraping and cleaning SNOTEL data and tutorials for data retrieval and training/modifying the neural models, are available at \url{https://clima.github.io/ClimaLand.jl/dev/generated/standalone/Snow/base_tutorial/}. CSVs of the training/testing data are also available here, as a quality-controlled fully observational dataset of physical variables for calibration and ML applications.


%



\appendix[A] 


\appendixtitle{Threshold Constraint Layers}

%



\subsection{Defining Threshold Constraint Layers}\label{a:threshold_layer}
\begin{figure*}[ht!]
        
        \centerline{\includegraphics[width = 39pc]{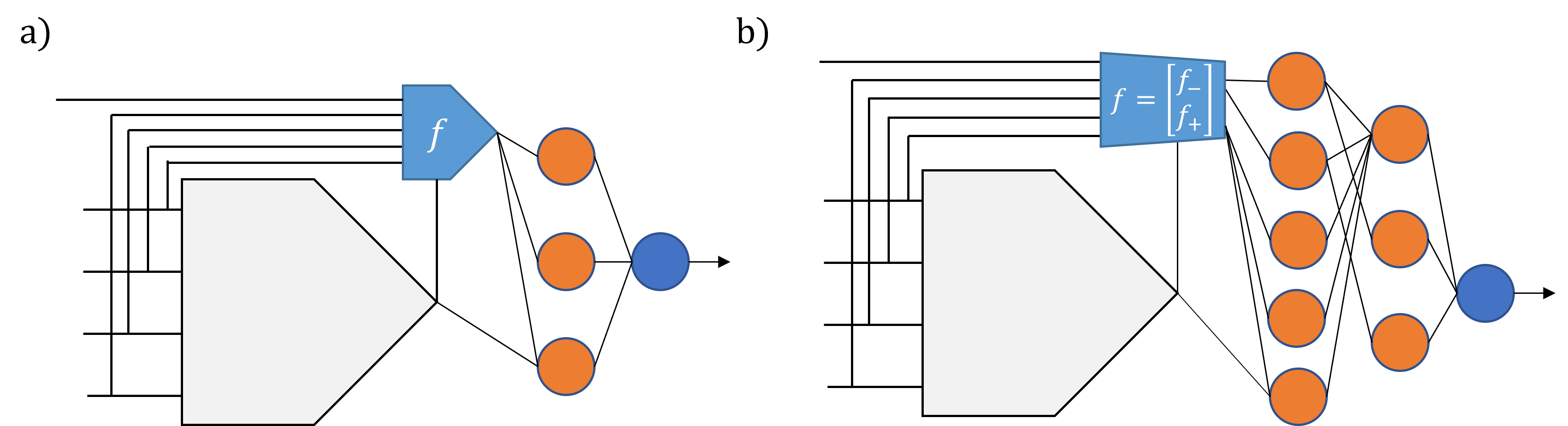}}
        \caption{In all graphics, inputs are on the left, with predictions on the right, and the grey pentagon depicts the predictive component from Fig.\ \ref{f:pred}. Knowledge of function $f$ can simplify layers by removing certain orange (ReLU) nodes, with resonant structures possible from the $\max, \min$ symmetry. Black weights are fixed at +1 or -1, with no biases or training. (a) General structure for a one-sided constraint (max or min) on the prediction. (b) General structure for a two-sided constraint (enforced range) on the prediction. }
        \label{f:structs}
        
\end{figure*}

Since $\mathrm{ReLU(x)} = \max(x, 0)$, we can re-express the minimum and maximum functions as
\begin{multline}
    \max(x, y) = y + \mathrm{ReLU}\left(x-y\right) = \\ \mathrm{ReLU}\left(y\right) - \mathrm{ReLU}\left(-y\right) + \mathrm{ReLU}\left(x-y\right) = \max(y, x),
\end{multline}
\begin{multline}
    \min(x, y) = y - \mathrm{ReLU}\left(y-x\right) = \\ \mathrm{ReLU}\left(y\right) - \mathrm{ReLU}\left(-y\right) - \mathrm{ReLU}\left(y-x\right) = \min(y, x).
\end{multline}
Then for a model output $p$ and any construction $f$ serving to threshold $p$, the bounds $\max(f, p)$ or $\min(f, p)$ can be explicitly implemented with a single depth-3 fixed-weight layer with no biases acting on input $[p, f]^\top$ with ReLU activation, followed by an accumulation with no activation:
\begin{multline} \label{eq:maxormin}
\begin{bmatrix}
\pm 1 & 1 & -1 
\end{bmatrix} \times \mathrm{ReLU}
\left(
\begin{bmatrix}
\pm 1 & \mp 1 \\
 0 & 1 \\
0 & -1
\end{bmatrix} \times
\begin{bmatrix}
p \\
f
\end{bmatrix}
\right) \\ = \bm{\mathsf{A}}_{1\pm} ^\top \: \mathrm{ReLU}\left(\bm{\mathsf{A}}_{2\pm} ^\top
\begin{bmatrix}
p \\
f
\end{bmatrix}
\right),
\end{multline}
where $+$ indicates $\max(f, p)$ and $-$ indicates $\min(f,p)$, and the ReLU acts element-wise. Eq.\ \ref{eq:maxormin} offers a simple formulation, but the symmetry of the max and min functions permits resonant structures—different weights yielding the same output. If the signs of $p$ or $f$ are unknown, both $+f$ and $-f$ (or $+p$ and $-p$) must be passed through the ReLU, along with $p-f$, to preserve all necessary information. However, if $f$ is always nonnegative (or nonpositive), the layer can be simplified from three layers to two, as passing $-f$ (or $+f$) becomes redundant and its ReLU always evaluates to zero. Similarly, if the threshold obeys $f \geq C$ (or $p \geq C$) for some constant $C$, a similar reduction is possible by including a bias term along with $A_{2\pm}$ and $A_{1\pm}$. The same reductions apply if $p$ also exhibits similar properties. Fig.\ \ref{f:structs}a illustrates the generalized structure for one-sided threshold constraints on a predictive component discussed in section 2\ref{s:structure}. 

Likewise, for a simultaneous upper bound $f_+$ and lower bound $f_-$ on $p$ for any constructions $f_+, f_-$ satisfying $f_+ \geq f_-$, we have
\begin{multline}
    \max(\min(p, f_+), f_-) = \\\mathrm{ReLU}(f_-) - \mathrm{ReLU}(-f_-) + \mathrm{ReLU}(\alpha),
\end{multline}
where
\begin{multline}
        \alpha = \mathrm{ReLU}(f_+) - \mathrm{ReLU}(-f_+) - \mathrm{ReLU}(f_-) + \\ \mathrm{ReLU}(-f_-) - \mathrm{ReLU}(f_+ - p),
\end{multline}
so the threshold can be explicitly implemented with a sequence of two fixed-weight layers containing no biases acting on input $[p, f_+, f_-]^\top$, followed by an accumulation with no activation:
\begin{multline} \label{eq:minmax}
\begin{bmatrix}
1 & 1 & -1 
\end{bmatrix} \times \mathrm{ReLU} \left(
\begin{bmatrix}
-1 & 1 & -1 & -1 & 1 \\ 
 0 & 0 & 0 & 1 & 0 \\ 
0 & 0 & 0 & 0 & 1 
\end{bmatrix} \right. \\
\left. \times \; \mathrm{ReLU} \left(
\begin{bmatrix}
-1 & 1 & 0 \\ 
0 & 1 & 0 \\  
0 & -1 & 0 \\ 
0 & 0 & 1 \\  
0 & 0 & -1    
\end{bmatrix} \times
\begin{bmatrix}
p \\
f_+ \\
f_-
\end{bmatrix}
\right) \right)
\end{multline}

\begin{equation}
    = \bm{\mathsf{A}}_{1+} ^\top \: \mathrm{ReLU}\left(\bm{\mathsf{A}}_{3} ^\top \mathrm{ReLU} \left(\bm{\mathsf{A}}_4^\top
\begin{bmatrix}
p \\
f_+ \\
f_-
\end{bmatrix}
\right) \right),
\end{equation}
which takes advantage of the identity $\mathrm{ReLU}(\mathrm{ReLU}(x)) = \mathrm{ReLU}(x)$. Like the one-sided threshold example, many resonant structures exist according to symmetry, and bounds on the thresholds or $p$ permit layer reductions. Fig.\ \ref{f:structs}b shows the generalized structure for a two-sided threshold constraint function $f$ outputting $f_+, f_-$ on the predictive component given in section 2\ref{s:structure}.

These constraint structures can adapt to any functional constraint $f$ of any input (including those independent of predictive inputs) or even the predictive component output $p$. This versatile framework exhibits minimal computational overhead without increasing runtime complexity. This thresholds an output rather than imposing an invariance or zeroing of a derivative, but this constraint class is relevant to many physical or nonnatural systems. This provides a simple means for emulating many systems when paired with universally approximating predictive models, or integrating data-driven parameterizations into larger models while enforcing constraints like physics or conservation laws. Absolute boundaries also enhance stability in time-stepping by keeping outputs realistic. 

Many code packages and languages cannot compute gradients for logical branches or min/max functions. This structure bypasses that limitation, allowing constraints to be enforced during training, even on legacy systems with minimal functionality. This mitigates the need to penalize the loss function, which can impede learning of the main objective \citep{tradeoff}. Enforcing constraints during training enables gradients and weight updates better-suited for predicting values within the boundaries. Unlike other penalty-free approaches like projecting outputs into constrained spaces, this method does not require constant or predefined thresholds and can explicitly predict boundary values rather than asymptotically approach them, which can inhibit learning by reducing gradient magnitudes \citep{quench}. Constraints under this construction can be analytically defined, or even parameterized and ``learned” from data during training, even simultaneously with the predictive component (e.g., a network for prediction and a network for boundary values, trained simultaneously), predicting both thresholds and values for entirely data-driven modeling. These functional forms can also be combined and stacked in larger networks or applied to non-network models. Figure \ref{f:structs} depicts the threshold constraint layers enveloping the predictive component from section 2\ref{s:structure}, but they could be placed inside larger networks, layered, or stacked as part of a larger predictive model, or applied on any non-network model. 

The primary benefit of this framework is enabling analytically prescribed bounds for trainable models during training. However, by representing the ``if over, then change" condition as a vectorized algebraic operation rather than logical branches, it eliminates branch misprediction in compilers and maximizes scalability for broadcasting and parallelization at large scale. This has performance implications for CPUs (see Table \ref{t:benchmarking}) but especially for GPUs with different (or no) protocols for branching \citep{GPUs}. When inputs for both $p$ and $f$ are similar, constraints can be implemented with one skip connection, streamlining the architecture. These constraint structures could be equivalently expressed using Maxout \citep{Maxout} or nested networks for a given constraint, but maintaining a single-network form with one skip connections results in faster training and sufficient variety in constraint expression. 

In the bigger picture, the framework’s adaptability and guarantee of constraints makes it a strong candidate for emulating systems where complex processes defy full analytical modeling but are bound by defined limits. This versatility opens doors for use-cases even beyond snow modeling. For instance, in two-sided threshold scenarios, $M$ can be viewed as interpolating between two boundaries, which we anticipate as offering utility in areas like predicting drag between turbulent and viscous limits or, transport in superdiffusive regimes. The approach holds potential to contribute to a comprehensive understanding of global dynamics, opening exciting avenues for future investigation. 

\appendix[B]
\appendixtitle{Data Methods}

Snow data comes from ground stations, aerial lidar surveys, or satellite images calibrated with ground measurements \citep{smyth-siteval}. Remote sensing offers broader coverage crucial for global modeling but can suffer obfuscation and lack the resolution to capture localized snow processes. Ground sensors can capture local effects but face challenges like malfunctions, terrain biases, and limited coverage, and aerial surveys track fewer variables. The SNOTEL network, with over 900 sites, uses automated sensors like snow pillows for SWE, ultrasonic sensors for depth, and weather instruments, transmitting data via ionospheric radio reflection \citep{meteorburst}. 

The SNOTEL network is rare in providing simultaneous collocated hydrometeorological data across varied climates, but faces notable data uncertainties like precipitation gauge undercatch or unphysical sensor values \citep{rasmussen, Hill19}. Not all sites measure all variables \citep{raleigh}, with many in wind-shielded or flat terrain, limiting climate diversity for ``universal” model calibration. All sites measure $\mathrm{SWE}$, $z$, and precipitation, but availability of variables like $T_\mathrm{snow}$, $T_\mathrm{soil}$, $R$, $\varphi$, and $v$ vary. Removing $T_\mathrm{soil}$ (due to its correlation with $\mathrm{ReLU}(T_\mathrm{air})$) and $T_\mathrm{snow}$ increased the usable sites to 44 in the continental U.S. and 7 in Alaska post-processing. 

Established snow science evaluation sites like Col De Porte \citep{coldeportdata}, K{\"u}htai \citep{khutai_data}, Reynolds Mountain East \citep{rme_data}, and Sodankyla \citep{sodankyla} were tested alongside Alaska SNOTEL data, which biases the evaluation towards lower elevations and similar $\bar{z}_+$. To address this, data from Yala Basecamp \citep{yala1, yala2} and two Rofental sites \citep{rofental} were included, though additional daily observational timeseries for all variables are presently minimal. Non-collocated measurements of snow depth ($z$), snow water equivalent ($\mathrm{SWE}$), or snow density ($\rho_{\mathrm{snow}}$) can lead to inconsistencies and biases in bulk density estimates. Training on such inconsistencies could benefit accuracy in large-scale climate modeling with gridded data, as it better matches the anticipated variability from coarse-graining, but it can also introduce validation biases if observation protocols differ from the training data. Therefore, all training and testing sites were assessed for the collocation of $z$ and SWE data collection using literature, imagery like \cite{smyth-siteval}, or direct communication with site representatives. This limited density evaluations to K{\"u}htai and CONUS/Alaskan SNOTEL sites.

Phenomena like wind drift can confound snow pillow data, as strong winds away from the wind sensor can push snow onto the pillow, creating positive measured $dz/dt$ with no precipitation and insufficient wind speeds for drift, as observed by \cite{meyer2012} in some SNOTEL sites. Our constraints in section 2\ref{s:constraints} can be violated by such events. However, only 1.5\% of the training data and 2.2\% of the testing data showed $dz/dt > 0$ without precipitation, and established models like Crocus and SNOWPACK also do not account for positive growth due to wind redistribution when run standalone. These models instead focus on the compaction or erosion of snow by wind \citep{crocus2012-err, SNOWPACK}; destructive effects our model captures. Ongoing work aims to integrate redistribution effects into snow models, and as this paper serves to introduce a framework compared to existing parameterizations, we find accounting for such effects beyond the scope of this study, representing an exciting area for future research. We invite interested parties to revise the variables, thresholds, and datasets in this initial formulation to the benefit of the community, as we envision broader adoption of this proposal beyond our specific formulation and calibration. 

\subsection{Data Cleaning Procedures} 

Established methods exist for cleaning SNOTEL daily $\mathrm{SWE}$, $ T_{air}$, and precipitation values, beginning with \cite{serreze99} and extended by \cite{yan2018}, but little consensus exists for depth or meteorological variables, particularly those available at hourly frequencies. 

Raw hourly and daily timeseries for all input variables were retrieved from the NRCS database, covering all entries available up to 2024-02-02. Bounds were applied to each sensor based on physical limits and limits in the SNOTEL sensor handbook \citep{NEH622C2}, removing any violative data. All solar, humidity, and wind speed timeseries were manually inspected, and any suspect or unphysical periods were flagged (see the function \texttt{manual\_filter()} in the code repository for a list of suspect periods). The following steps were then taken in order per site: 

\begin{itemize} 

\item Weekly maximum hourly wind speeds were determined, and the median $\tilde{w_{max}}$  and interquartile range IQR were calculated. Observed wind values $w_i$ were flagged if $\frac{w_i - \tilde{w_{max}}}{\mathrm{IQR}} > 6$. For timeseries with over 24 flagged values, ``blocks” of flagged values were grown in 72-hour steps until no further flags existed. If more than 5\% of values in a block were flagged, all observations in that block were flagged. 

\item Unflagged hourly wind speed, solar radiation, relative humidity, and air temperature observations were binned into two-week windows and by hour, with the mean calculated for each hourly bin to generate an annual profile (24 hours per 26 biweeks). Gaps of 6 hours or less were filled using linear interpolation, while gaps of 6 to 24 hours were filled using the appropriate profile.

\item For hourly $z$ data, for each non-missing/flagged observation $z_i$, the values $dz_+ = z_{i+1}-z_i$, $dz_- = z_{i}-z_{i-1}$, and $dt_+ = t_{z_{i+1}}-t_{z_i}$ were calculated. A threshold $Z = 20$ inches was picked and $z_i$ having $dz_- \geq Z$ and $dz_+ \leq -Z$, or  $dz_- \leq -Z$ and $dz_+ \geq -Z$ were flagged. A ``rut” of bad data began when $dz_- \geq Z$ and $|dz_+| \leq Z$, and continued until $|dz_+| \geq Z$, all observations in a rut were flagged. A rut lasting more than 20 observations or $dt_+ > 30$ days resulted in all observations being flagged until $z_i=0$. From April through August, after the first time reaching $z_i =0$, nonzero $z_i$ were flagged. All $z_i$ having $\frac{z_i}{\mathrm{SWE_i}}$ less than 1, over 50, or missing $\mathrm{SWE_i}$ were flagged. This procedure was iterated until no more $z_i$ were flagged, and was designed based on the structure of sensor errors in the hourly $z$ data. This was only for the hourly data, as daily SNOTEL depth data are quality checked. 

\item All hourly timeseries were then binned into three 8-hour bins per day. For solar radiation, wind speed, air temperature, and humidity, the mean of all non-missing/flagged values per bin determined (if all were missing, a missing value was given). These three averages were averaged to create the daily values, and a daily value reported as missing if any 8-hour bin average was missing. For hourly $z$, the day’s value was the first available observation, or missing if all observations that day were missing. 

\item The annual maximums of all generated daily solar radiation observations were determined, and the median $\tilde{R_{max}}$ and interquartile range IQR of these maximums determined, and all daily solar values $R_i$ with a score $\frac{R_i - \tilde{R_{max}}}{\mathrm{IQR}} > 2$ were excised. This only removed a handful of individual irregular spikes in the rolled-up data and left most sites unaltered. 

\item The converted daily timeseries was coalesced with the daily raw timeseries to form a complete daily timeseries for the site. Raw daily $z$ and air temperature values took priority over converted-hourly values if both values existed, and converted-hourly values for solar radiation, relative humidity, wind speed took priority over daily values if both existed (only the raw daily data was used for $\mathrm{SWE}$ and accumulated precipitation). 

\item Air temperature corrections to the air temperature data were applied in accordance with \cite{snotel-temp} and associated metadata of which sites to correct, as of May 2024. 

\item Standard quality control procedures and flagging for $\mathrm{SWE}$, accumulated precipitation values, and air temperature as given in \cite{serreze99} and extended by \cite{yan2018} were implemented. Inconsistent water years with maximum $\mathrm{SWE}$ at least 5\% greater than the associated accumulated precipitation value were excised in accordance with this protocol.

\item Remaining accumulated precipitation observations were edited to account for gauge undercatch, following the procedure outlined in \cite{livneh} used in \cite{yan2019}. 

\item Daily timeseries of $z$ were compared to the quality controlled $\mathrm{SWE}$, and any values showing $\frac{z_i}{\mathrm{SWE_i}}$ less than 1, over 50, or missing $\mathrm{SWE_i}$ were excised. 

\item For all variables, gaps of 3 days or less were then filled via linear interpolation. 

\item All data was then scaled into SI units ($z$, $\mathrm{SWE}$, and accumulated precipitation to meters from inches, relative humidity scaled from 0-1, and wind speed to meters per second from kilometers per hour) 

\item Only days with complete cases (no missing values in all variables) were extracted, and sequential differences $\Delta z_i = z_{i+1} - z_i$, $\Delta \mathrm{SWE}_i = \mathrm{SWE}_{i+1}-\mathrm{SWE_i}$, $p_i = AP_{i+1} - AP_i$ were calculated, where $AP$ is accumulated precipitation. Only values where $\Delta t_i  = t_{i+1}-t_i = 1$ day were kept, and the target $\frac{dz}{dt}_i \approx \frac{\Delta z_i}{\Delta t_i}$ created, as well as analogous $\frac{d \mathrm{SWE}}{dt}_i$ and $P_i = p_i/\Delta t_i$. Days with $P_i < 0$ due to resetting of the water year were changed to $P_i$ = 0. 

\end{itemize} 

The data at this point was saved. Upon importing for usage in model training, precipitation $P_i$ was split into snow $P_\mathrm{snow}$ and rain $P_\mathrm{rain}$ based off temperature and humidity values using the snow fraction equation from \cite{rain-snow-split} with over an 88\% success rate across the northern hemisphere: 

\begin{equation}
    f_{\mathrm{snow}} = \frac{1}{1+e^{\alpha + \beta T_{\mathrm{air}} + \gamma \varphi}},
\end{equation}
with $\alpha = -10.04$, $\beta = 1.41 \; ^\circ \mathrm{C}^{-1}$, and $\gamma = 9$ (with the relative humidity $\varphi \in [0,1]$). The correlation of $P_\mathrm{rain}$ data after quality control procedures and feature engineering to $dz/dt$ and $d\mathrm{SWE}/dt$ was the lowest of all variables among the training data at r = -0.016 and 0.005, respectively, informing the choice to also remove it from the dataset to further increase computational simplicity and scalability of the final model. Reintroduction and retraining including the variable on the optimized model structure to verify this choice resulted in negligible changes in performance at the site level and on average. 

Three exceptions apply to the above procedure specifically for this set of training data, which might not apply for other SNOTEL sites, and are as follows: 

\begin{itemize} 

\item Air temperatures in the continental US were bounded below by $-40^{\circ}$C and air temperatures in Alaska were bounded below by $-50^{\circ}$C instead of the instrument limit of $-60^{\circ}$C to remove on average 1-2 individual suspect temperature spikes per site. 

\item All raw $z_i >$ 175 inches were flagged (a bound solely for removing unphysical sensor spikes in this specific training data, and should be checked for alternative data) 

\item For SNOTEL site 1122, the averaged air temperature hourly timeseries took priority over the daily timeseries when coalescing data. 

\end{itemize} 

The code to scrape SNOTEL data from the NCRS database as and apply the above processing as well as a tutorial has been made publicly available in the code repository.

The other sites were provided varying degrees of quality control beyond unit conversion and generation of targets $dz/dt$ and $d\mathrm{SWE}/dt$ from the resulting data. For Col De Porte, K{\"u}htai, and Reynolds Mountain East, no action was taken beyond collecting data into daily timeseries (averaging all data except taking the first available $z$, $\mathrm{SWE}$ measurement per day, no need for intermediate 8-hour blocks). Precipitation was also split into rain and snow following the same procedure as SNOTEL data for K{\"u}htai. In the Rofental's Bella Vista site data, a nonzero offset of precipitation data was subtracted from dates after 2022-01-01, and otherwise both sites were collected to daily data directly and treated with the same precipitation undercatch procedures as the SNOTEL data. At the Yala Basecamp site, only the 2018 year was taken due to feature availability. All negative $\mathrm{SWE}$, $z$ values were set to zero, and gaps in the $\mathrm{SWE}$ data up to 9 consecutive values were filled with linear interpolation. The same undercatch procedures as SNOTEL were provided, and data was aggregated to the day level directly (for $z$, the median was taken to ignore sensor spikes, otherwise the mean for all variables). For Sodankyla, the following measures were taken from the raw data series beyond unit conversion, target feature creation in the same manner as the SNOTEL sites, and direct aggregation to daily level:

\begin{itemize}
    \item 1-minute $\mathrm{SWE}$ data was aggregated to the 10-minute level (the same level as other variables).
    \item All 10-minute variables had gaps up to 9 consecutive values filled with linear interpolation.
    \item All solar radiation data less than zero was set to zero.
    \item Missing $z$ data from May to November 2016 was set to zero.
    \item The same undercatch procedure was applied to precipitation data as the SNOTEL data.
\end{itemize}

\appendix[C]
\appendixtitle{Hyperparameters and Model Benchmarking}
\section{Hyperparameters}
Optimal hyperparameters are summarized in Table \ref{t:hyperparam}. Scores were evaluated using 44-fold leave-one-out cross-validation with a batch size of 64, tracking performance every 10 epochs over 200 epochs. Most timeseries trials performed optimally when training for approximately 100 epochs. A nonzero value of $n_2$  emphasized extreme points in the custom loss function, enhancing accumulated predictions, particularly for datasets with few extreme samples. The optimal hyperparameters yielded a network size of 435 trainable parameters for the $z$ network and 540 for the SWE network, compared to around 50 empirically tuned constants (parameters and internal code) in Snow17 for predicting SWE and $z$. 
\begin{table*}[ht!]
\caption{Hyperparameter results. For timeseries, NSE and RMSE were the primary metrics, while RMSE was the main metric for regression. $n_1$ = 2 (L2-like metric) provided the lowest RMSE for all choices, which is unsurprising. However, it is interesting that $n_1$ = 2 also minimized SPE (an L1-like metric) compared to $n_1$ = 1.}
\label{t:hyperparam}
\begin{center}
\begin{tabular}{cccccc}
\topline
Parameter & Description & Range & Series Score: $z$ & Series Score: $\mathrm{SWE}$ & Regression Score: $dz/dt$\\ \midline
$N$ & Averaging Consecutive $N$ Days & 1, 2, 3 & 1 & 1 & 1\\
$n$ & Width of Mixing Layer & 3, 4, 5, 6 & 4 & 5 & 6\\
$n_1$ & Power Scaling of Prediction Error & 1, 2 & 2 & 2 & 2\\
$n_2$ & Power Scaling of Target Magnitude & 0, 0.1, 1, 2, 4 & 4 & 2 & 2\\
\botline
\end{tabular}
\end{center}
\end{table*}

\section{Model Benchmarking}
Table \ref{t:performance} presents the means and medians of different model configurations for timeseries generation at the testing sites. The neural parameterization is also compared to another network, $M_\mathrm{ifelse}$, which has identical predictive structure but has no boundaries during training and only calculates and applies thresholds post-training through vectorized if/else logic.
The regression RMSE scores between $M$ and $M_\mathrm{ifelse}$ on training and testing data were within 1-2 mm/day, with $M_\mathrm{ifelse}$ performing better on training regression, despite notable superiority in $M$ for timeseries accuracy. This trend of better training regression for $M_\mathrm{ifelse}$ but worse timeseries generation persisted across repeated training trials, reinforcing the notion that incorporating bounds during training enhances physical representation and generalizability, rather than minimizing training loss at the expense of other beneficial properties.

\begin{table*}[ht!]
\caption{Performance of the models (for SWE and $z$) and parameterizations (for $z$) in this study across testing sites, using labels from section 2\ref{s:testing}. Medians are listed, with the mean in parenthesis alongside the median. Another network $M_\mathrm{ifelse}$ has been included, which trains without constraints and applies them during testing with if/else statements, to highlight the performance gains over out-of-sample data from including the framework developed in this paper. Subscripts on metrics indicate whether depth or SWE was benchmarked.}
\label{t:performance}
\begin{center}
\begin{tabular}{cccccc}
\topline
Metric          & \multicolumn{2}{c|}{Standalone Model (SWE, $z$)}     & \multicolumn{3}{c}{Parameterization ($z$)}          \\
               & $\tilde{M}$            & \multicolumn{1}{c|}{SN17} & $M$             & SN17O         & $M_\mathrm{ifelse}$    \\
\midline
$\mathrm{MAE}_z$ (cm)    & 6.9 (7.8)     & 8.2 (16.1)                & 4.6 (5.9)     & 6.9 (8.9)     & 6.4 (8.0)     \\
$\mathrm{RMSE}_z$ (cm)   & 11.3 (12.9)   & 12.9 (23.7)               & 8.3 (9.6)     & 10.1 (14.3)   & 10.6 (13.0)   \\
$\mathrm{NSE}_z$         & 0.915 (0.870) & 0.914 (0.347)             & 0.955 (0.936) & 0.937 (0.782) & 0.925 (0.875) \\
$\mathrm{SPE}_z$ (\%)    & 11.9 (13.4)   & 11.5 (27.6)               & 8.8 (9.6)     & 9.8 (15.4)    & 11.5 (13.5)   \\
$\mathrm{MAE}_\mathrm{SWE}$ (cm)  & 2.1 (2.3)     & 2.5 (4.2)                 & -             & -             & -             \\
$\mathrm{RMSE}_\mathrm{SWE}$ (cm) & 3.7 (4.1)     & 4.4 (7.4)                 & -             & -             & -             \\
$\mathrm{NSE}_\mathrm{SWE}$       & 0.931 (0.862) & 0.933 (0.663)             & -             & -             & -             \\
$\mathrm{SPE}_\mathrm{SWE}$ (\%)  & 11.3 (14.2)   & 11.3 (21.4)               & -             & -             & -            \\
\botline
\end{tabular}
\end{center}
\end{table*}

Table \ref{t:significance} lists the p-values from Wilcoxon Signed Rank tests used in this study. This non-parametric test assesses the significance of differences between matched samples (timeseries RMSE, in this case), comparing the performance of the presented framework over Snow17 and $M_\mathrm{ifelse}$. The lack of significance for $M$ at validation sites was expected, as all models were calibrated for performance on this data. However, significant improvements in out-of-sample testing data highlight the advantages of the presented approach. $\tilde{M}$'s significant difference from Snow17 in validation sites was unexpected, potentially due to Snow17's calibration prioritizing SWE directly (one variable), while $\tilde{M}$ benefits from optimizing $z$ and SWE as direct inputs for SWE prediction.

\begin{table*}[ht!]
\caption{Statistical significance testing of model RMSE at validation and testing sites. p-values of the Wilcoxon Signed Rank test are shown, which compares the predictive power of two models. Labels follow those in Table \ref{t:performance}. Significant ($p<0.05$) values are highlighted in bold. $M_\mathrm{ifelse}$ has been added to underscore the significant improvement on out-of-sample data when applying the demonstrated framework.}
\label{t:significance}
\begin{center}
\begin{tabular}{ccccc}
\topline
Site RMSE             & $M$ vs. SN17O ($z$)     & $\tilde{M}$ vs. SN17 ($z$)  & $\tilde{M}$ vs. SN17 (SWE) & $M$ vs. $M_\mathrm{ifelse}$ ($z$) \\
\midline
Validation Sites & 0.673          & \textbf{0.006} & \textbf{0.029}  & 0.903           \\
Testing Sites    & \textbf{0.013} & 0.135          & 0.268           & \textbf{0.0003} \\
\botline
\end{tabular}
\end{center}
\end{table*}

Table \ref{t:benchmarking} presents time/memory benchmarking. Testing was conducted on one Intel i9 CPU (no GPU). Models were tested in ``Column" mode, processing one location's inputs at a time (like a site simulation), and in ``1.5M Grid" and ``150M Grid" modes, evaluating input vectors of $\sim$1.5 million and $\sim$150 million locations’ inputs at once (like a global model at 10km or 1km land resolution), by stacking 14 or 141 copies of all SNOTEL inputs, respectively. Average memory/time for a single evaluation (excluding garbage collection and compilation) were tracked and normalized by the number of inputs. Column mode was averaged over 10 SNOTEL data passes (10.6 million trials), while 1.5M and 150M Grid results were averaged over 250 trials. Snow17 can only iterate between locations in a Column-like mode. Both $M$ and $M_\mathrm{ifelse}$ determine boundaries and adjust outputs accordingly, but $M$ does this structurally, while $M_\mathrm{ifelse}$  uses vectorized boundary creation and broadcasts conditional if/else logic, creating the opportunity for branch misprediction effects.

While $M_\mathrm{ifelse}$  is slightly faster with less memory in Column mode  (comparing and adjusting one value against two values is quicker than processing three values through matrix multiplication),  $M$ is faster by about 10\% over gridded inputs (0.147 and 1.47 seconds total, respectively). Snow17 with data assimilation requires more time and memory than without, as assimilation occurs after the primary evaluation. $\tilde{M}$ is larger than standalone Snow17 (and about twice that of $M$, as expected) but evaluates faster, which benefits long simulations across many locations in global models. However, $M$ is quicker and requires less memory than Snow17. Results may vary based on programming languages, libraries, compilers, or protocols used for implementation, but one could always train utilizing the framework (the main benefit) and alter boundary implementation afterwards as desired. We anticipate these findings would extend to GPUs that often lack optimized branching, suggesting further advantages for large-scale model integration. Additional comprehensive benchmarking remains an avenue for future research. 

\begin{table*}[ht!]
\caption{Time and memory benchmarking of all models (for $z$ and SWE) and parameterizations (for $z$), listing required resources per evaluation for single-instances (``Column'') or per-instance over roughly 1.5 million or 150 million instances (``1.5M Grid'' and ``150M Grid'') simultaneously. Snow17 only evaluates single instances, and $M_\mathrm{ifelse}$ is listed to compare constraint layers against vectorized if/else post-processing. All benchmarks were evaluated on a single Intel i9 CPU.}
\label{t:benchmarking}
\begin{center}
\begin{tabular}{cccccc}
\topline
Metric                 & \multicolumn{2}{c|}{Standalone Model ($z$, SWE)} & \multicolumn{3}{c}{Parameterization ($z$)} \\
                       & $\tilde{M}$                & \multicolumn{1}{c|}{SN17}             & $M$        & SN17O     & $M_\mathrm{ifelse}$    \\
\midline
Column, $T$ ($\mu$s)      & 2.6              & 3.6             & 1.3     & 3.8      & 1.0          \\
Column, Allocated Memory (KB)  & 2.5              & 1.6             & 1.2     & 1.7       & 1.1          \\
1.5M Grid, $T$ ($\mu$s)     & 0.21              & -                & 0.099     & -         & 0.11          \\
1.5M Grid, Allocated Memory (KB) & 0.80              & -                & 0.37     & -         & 0.34            \\
150M Grid, $T$ ($\mu$s)     & 0.21              & -                & 0.099     & -         & 0.11          \\
150M Grid, Allocated Memory (KB) & 0.80              & -                & 0.37     & -         & 0.34            \\

\botline
\end{tabular}
\end{center}
\end{table*}

\newpage

\bibliographystyle{ametsocV6}
\bibliography{main}

\end{document}